\pdfoutput=1
\documentclass[10pt,twocolumn,letterpaper]{article}

\usepackage[pagenumbers]{iccv} % To force page numbers, e.g. for an arXiv version

\usepackage{amsmath,amsfonts,bm}

\def\eqref#1{equation~\ref{#1}}
\def\1{\bm{1}}

\def\rmA{{\mathbf{A}}}

\def\rmD{{\mathbf{D}}}

\DeclareMathAlphabet{\mathsfit}{\encodingdefault}{\sfdefault}{m}{sl}
\SetMathAlphabet{\mathsfit}{bold}{\encodingdefault}{\sfdefault}{bx}{n}

\usepackage{graphicx}
\usepackage{subcaption}
\usepackage{float}
\usepackage{lscape}                                         % Useful for wide tables or figures.
\usepackage{wrapfig}

\usepackage[lined,ruled,linesnumbered]{algorithm2e}
\usepackage{animate} %% convert frames to video

\usepackage{booktabs}                   % Publication quality tables
\usepackage{multirow}

\usepackage{makecell}

\usepackage{paralist}
\usepackage{bm}                          % Make bold, italic math symbols
\usepackage{epsfig}                      % for figures
\usepackage{mathtools}
\usepackage{amssymb,amsmath}   % Short math guide for LaTeX ftp://ftp.ams.org/pub/tex/doc/amsmath/short-math-guide.pdf

\usepackage{units}
\usepackage{color}

\usepackage{comment}

\usepackage{xspace}
\usepackage{setspace}
\usepackage{grfext}
\PrependGraphicsExtensions*{.jpg,.png,.PNG}

\usepackage[symbol]{footmisc}

\usepackage{soul} % \ul
\usepackage{multirow}

\usepackage{pifont}
\def\xi{\mathbf{x}_i}

\graphicspath{{figures}, {examples}}

\RequirePackage[shortlabels,inline]{enumitem}
\setlist[itemize]{noitemsep,leftmargin=*,topsep=0em}
\setlist[enumerate]{noitemsep,leftmargin=*,topsep=0em}

\newcommand{\mpage}[2]
{
\begin{minipage}{#1\linewidth}\centering
#2
\end{minipage}
}

\newcommand{\topic}[1]
{
\vspace{1mm}\noindent\textbf{#1}
}
\newcommand{\methodname}{{HALO}\xspace}

\definecolor{iccvblue}{rgb}{0.21,0.49,0.74}
\usepackage[pagebackref,breaklinks,colorlinks,allcolors=iccvblue]{hyperref}

\def\paperID{} % *** Enter the Paper ID here
\def\confName{ICCV}
\def\confYear{2025}

\title{HALO: Human-Aligned End-to-end \\ Image Retargeting with Layered Transformations}

\author{Yiran Xu$^{1,3*}$,
Siqi Xie$^{2}$,
Zhuofang Li$^{2}$,
Harris Shadmany$^{2*}$,
Yinxiao Li$^{1}$,
Luciano Sbaiz$^{1}$, \and 
Miaosen Wang$^{1}$,
Junjie Ke$^{1}$,
José Lezama$^{1}$,
Hang Qi$^{1}$,
Han Zhang$^{4*}$,
Jesse Berent$^{1}$, \and 
Ming-Hsuan Yang$^{1}$,
Irfan Essa$^{1}$,
Jia-Bin Huang$^{3}$,
Feng Yang$^{1}$ \vspace{3mm} \\
$^{1}$Google DeepMind, $^{2}$Google, $^{3}$University of Maryland, College Park,  $^{4}$Reve AI\\
\vspace{-5mm}
}

\begin{document}
\twocolumn[{
\renewcommand\twocolumn[1][]{#1}
\maketitle
\begin{center}
    \centering
    \captionsetup{type=figure}
    \includegraphics[trim=0 0 0 0, clip,width=\linewidth]{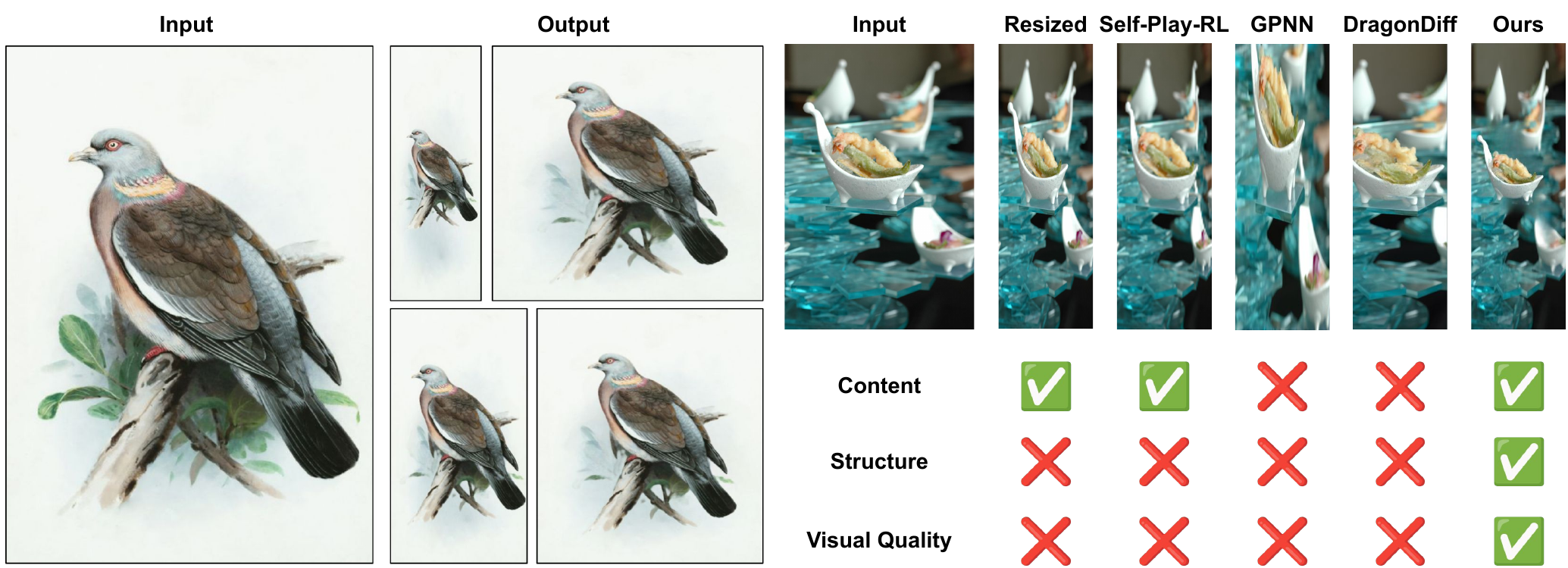}
    \captionof{figure}{
    \textbf{Content- and structure-aware image retargeting.}
    Our method, \methodname, takes a single image as input and reformats it for different aspect-ratios.
    Compared to previous methods: Self-Play-RL~\cite{kajiura2020self}, GPNN~\cite{granot2022drop}, and DragonDiffusion~\cite{mou2024dragondiffusion}, our method shows better performance in preserving the structure and content of the input image and has better visual quality. 
    % Once trained, our method generalizes to different categories of images, including animals, natural landscapes and buildings without \emph{finetuning}.
    % \methodname generalizes well to in-the-wild input~\cite{unsplash2020unsplash} to different target sizes without \emph{finetuning}.
    %We show images from~\cite{unsplash2020unsplash}.
    }
\end{center}
}]

\let\thefootnote\relax\footnotetext{$^{*}$Work done when Yiran, Harris, and Han worked at Google.}

\begin{abstract}

Image retargeting aims to change the aspect-ratio of an image while maintaining its content and structure with less visual artifacts. 
Existing methods still generate many artifacts or fail to maintain original content or structure.  
To address this, we introduce \methodname, an end-to-end trainable solution for image retargeting. 
%The core idea of \methodname is to warp the input image to target resolution. 
Since humans are more sensitive to distortions in salient areas than non-salient areas of an image, \methodname decomposes the input image into salient/non-salient layers and applies different wrapping fields to different layers. To further minimize the structure distortion in the output images, we propose perceptual structure similarity loss which measures the structure similarity between input and output images and aligns with human perception. Both quantitative results and a user study on the RetargetMe dataset show that \methodname achieves SOTA. 
Especially, our method achieves an 18.4\% higher user preference compared to the baselines on average.
% our method increases human preference by 11.6\% compared with the second best method.

% %A user study also shows a $43.84 \%$ preference for \methodname. 
%Our method shows significant improvement across content preservation and structural integrity over different evaluation metrics.

%To allow flexibly manipulation for different regions of the images according to their importance, \methodname decomposes the input image into salient/non-salient layers and applies distinct warping fields to transform each layer individually. Additionally, \methodname introduces a novel training paradigm that needs no paired training data.
%These novel ideas allows our model to produce distortion-free results while minimizing content loss.
%To validate our model, we conduct extensive experiments with existing baselines on the RetargetMe benchmark. 
%Our method shows significant improvement across content preservation and structural integrity over different evaluation metrics.
%A user study also shows a $43.84 \%$ preference for \methodname. 
%Furthermore, \methodname demonstrates strong generalization across diverse image types and target aspect-ratios.

\end{abstract}
% Story line
% 1. Retargeting is important...
%   Definition of retargeting: 
    % i)   Content (high-level)
    % ii)  Structure/shape/layout (low/mid-level)
    % iii) No visual artifacts.
% 2. Challenge i): No paired training data
    %    Previous methods use weakly supervision, but still poor (see Fig.2). 
    %    Why? 
%   Challenge ii): single warping field for the whole image -> lack of flexibility. 
    %  why? bg -> do not need too much aggressive transformation
    % fb -> need aggressive transformation
%   Challenge iii): using a generative model to model an image
% 3. Our paper
    %  i)   We propose a novel weakly-supervision training framework with random transformation disturb.
    %  ii)  We predict two different transformations for fg and bg separately.
    %  iii) The transformation is applied to the whole fg or whole bg -> No duplicated/missing objects. 
% 4. Our contributions
    %   i)  An novel training pipeline for image retargeting
    %  ii)  A feed-forward network for fast image retargeting (compared to generative models).
    %  iii) Outperform SoTAs in different metrics.

\section{Introduction}\label{sec:intro}
% Why we need image retargeting
% Images are displayed on a diverse set of platforms and devices, each with a different aspect-ratio. 
% %% Why don't these simple tricks work?
% Resizing or cropping images are traditional approaches for it, but resizing can distort structures, and cropping inevitably removes content.
% Image retargeting~\cite{rubinstein2010retargetme,tang2019retargetability} seeks to address these problems and adjusts an image's aspect-ratio while preserving its key content and structure.
% % Problem formulation
% %Given an input image and a target aspect-ratio, image retargeting aims to generate an output image that preserves the content and layout from the input.
% As defined by~\cite{rubinstein2010retargetme,vaquero2010survey}, a successful image retargeting outcome is as follows: 
Images are displayed across various platforms and devices with differing aspect ratios. Traditional resizing distorts structures, while cropping removes content. 
Image retargeting~\cite{rubinstein2010retargetme,tang2019retargetability} addresses this by adjusting aspect ratios while preserving key content and structure. As defined by~\cite{rubinstein2010retargetme,vaquero2010survey}, a successful retargeting should achieve the following criteria:
(a) Key \emph{content} in the input image should be preserved in the output image; 
(b) Inner \emph{structure} of the input should be maintained in the output;
(c) There should be \emph{no distortion} or \emph{visual artifacts} in the output image.
% \yl{This can be supported with a teaser figure. Add text after adding the teaser figure}
% \yiran{Added to the teaser. Thanks.}

\begin{figure*}[h!]
    \centering
    % \begin{subfigure}[t]{0.35\textwidth}
    %     \centering
    %     \frame{\includegraphics[trim=0 0 0  clip,height=0.12\textheight]{images/motivation_assets/orchid.png}}
    %     \caption{{Input}}
    % \end{subfigure} \hfill
    % \begin{subfigure}[t]{0.2\textwidth}
    %     \centering
    %     \frame{\includegraphics[trim=0 0 0 0, clip,height=0.12\textheight]{images/motivation_assets/orchid_0.50_qp.png}}
    %     \caption{{QP}}
    % \end{subfigure} \hfill
    % \begin{subfigure}[t]{0.2\textwidth}
    %     \centering
    %     \frame{\includegraphics[trim=0 0 0 0, clip,height=0.12\textheight]{images/motivation_assets/orchid_0.50_gpdm.png}}
    %     \caption{{GPDM}}
    % \end{subfigure} \hfill
    % \begin{subfigure}[t]{0.2\textwidth}
    %     \centering
    %     \frame{\includegraphics[trim=0 0 0 0, clip,height=0.12\textheight]{images/motivation_assets/orchid_0.50_selfplay.png}}
    %     \caption{Self-Play-RL}
    % \end{subfigure}
    
    \begin{subfigure}[t]{0.20\textwidth}
        \centering
        \frame{\includegraphics[trim=0 0 0 0, clip,width=\textwidth]{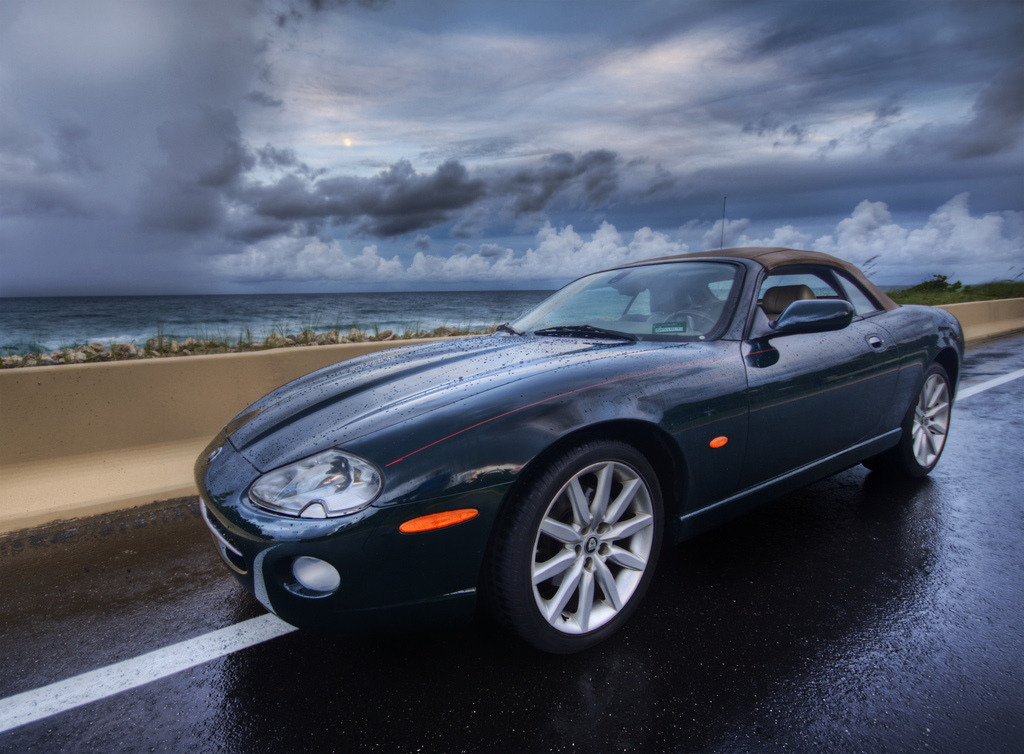}}
        \caption{{Input}}
    \end{subfigure} \hspace{-15mm} \hfill 
    \begin{subfigure}[t]{0.25\textwidth}
        \centering
        \frame{\includegraphics[trim=0 0 0 0, clip,width=\textwidth]{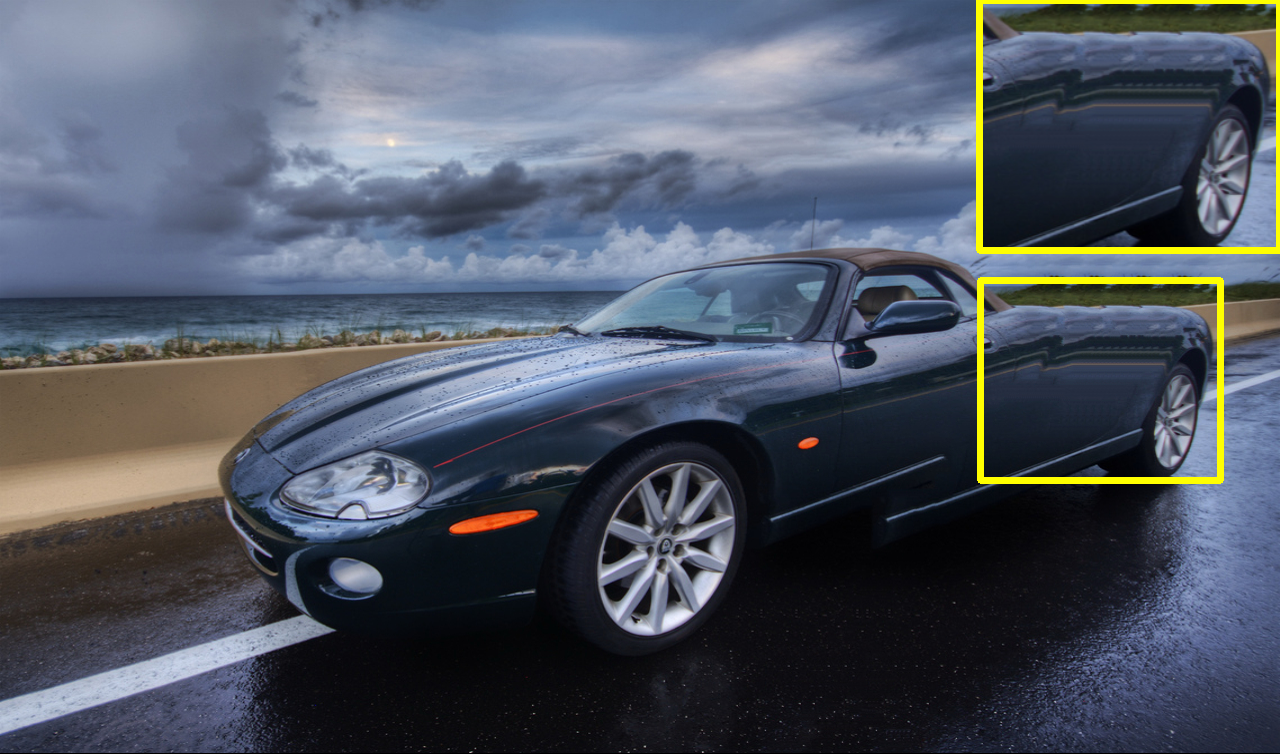}}
        \caption{{Shift-Map~\cite{pritch2009shift}}}
    \end{subfigure}  \hspace{-15mm} \hfill
    \begin{subfigure}[t]{0.25\textwidth}
        \centering
        \frame{\includegraphics[trim=0 0 0 0, clip,width=\textwidth]{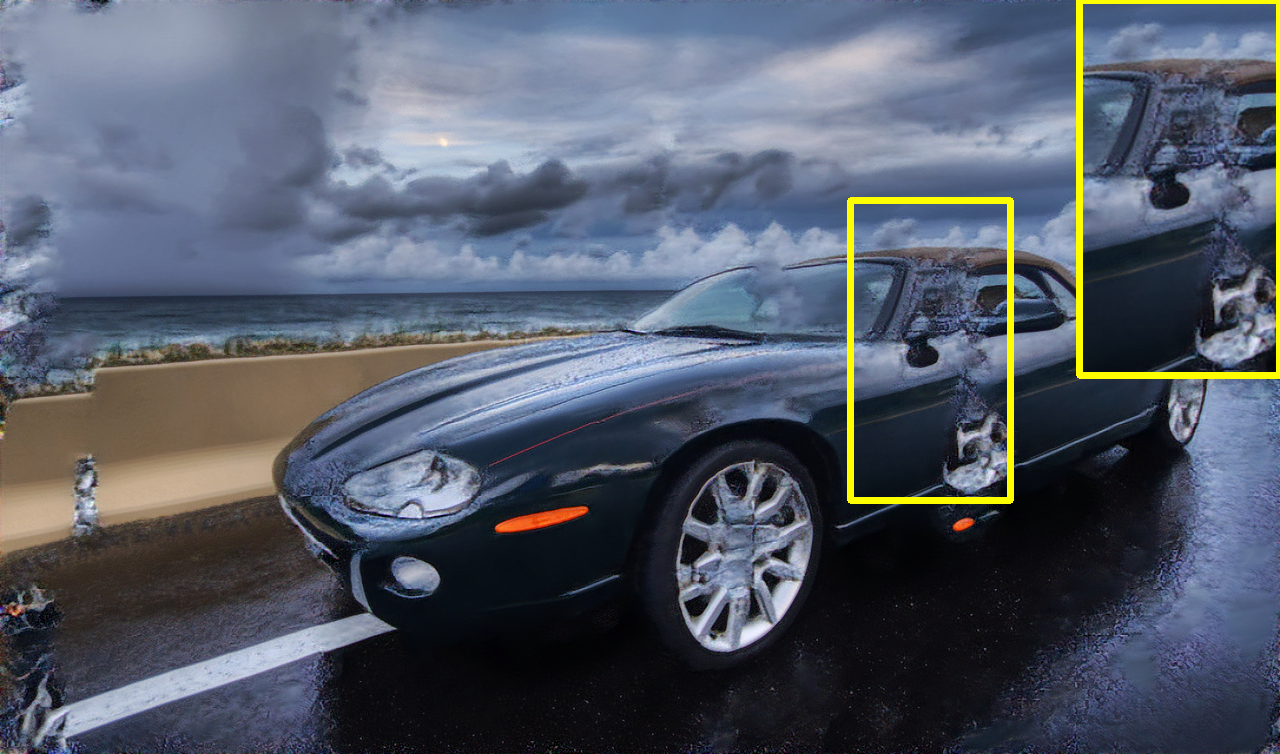}}
        \caption{{GPDM~\cite{cho2017wssdcnn}}}
    \end{subfigure} \hspace{-15mm} \hfill 
    \begin{subfigure}[t]{0.25\textwidth}
        \centering
        \frame{\includegraphics[trim=0 0 0 0, clip,width=\textwidth]{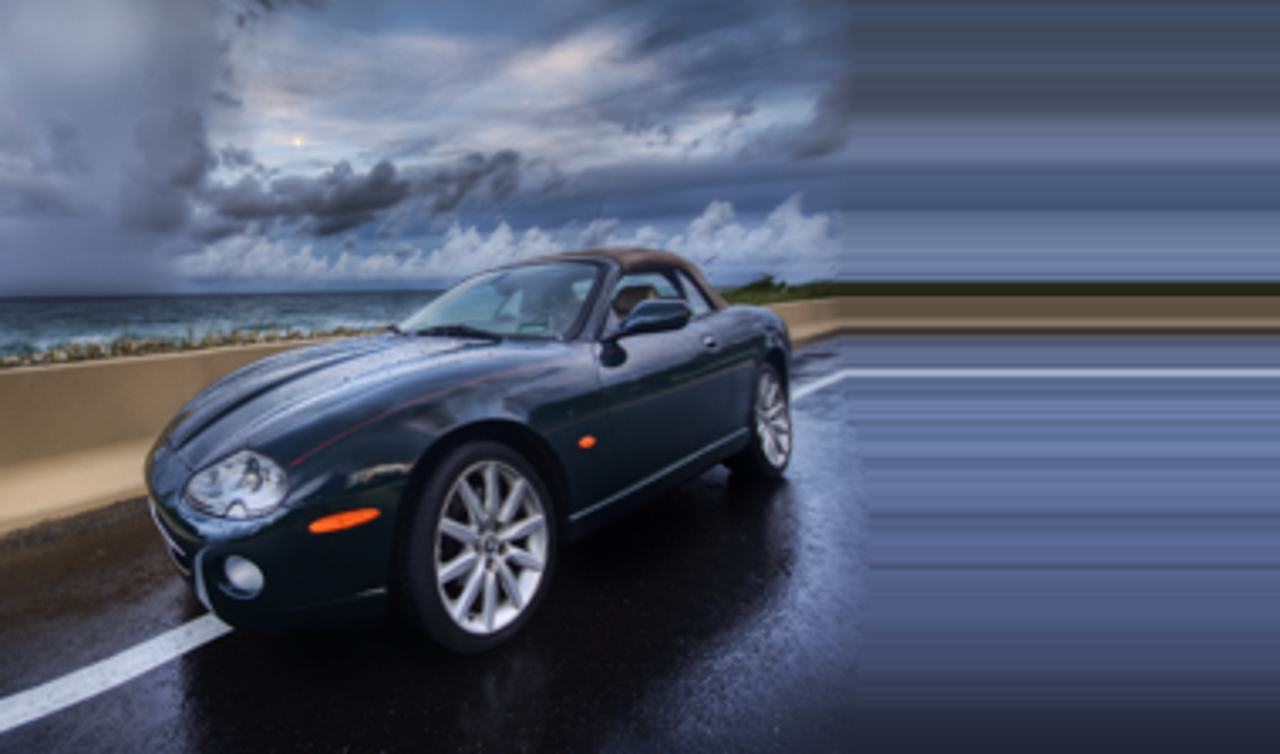}}
        \caption{WSSDCNN~\cite{cho2017wssdcnn}}
    \end{subfigure}
    
    \caption{\textbf{Limitations of exisiting retargeting methods.}
    Previous image retargeting methods have difficulty preserving the input image content and structure. 
    % They either drop the content or have distorted structure. 
    (b) A traditional method Shift-Map~\cite{pritch2009shift} duplicates the structure of the car. (c) A generative modeling method GPDM~\cite{elnekave2022generating} adds extra content. (d) A feed-forward method WSSDCNN~\cite{cho2017wssdcnn} introduces out-of-boundary (OOB) artifacts.}
    \label{fig:motivation}
\end{figure*}
% Challenges
Developing a good image retargeting solution is challenging. Particularly, there are no ground-truth image pairs for training. Researchers proposed many algorithms including traditional optimization approaches~\cite{liu2005automatic,setlur2005automatic,wolf2007non,simakov2008summarizing,rubinstein2009multi,barnes2009patchmatch,pritch2009shift,rubinstein2010retargetme,chen2010content,shi2010content}, reinforcement learning~\cite{kajiura2020self}, generative modeling methods~\cite{elnekave2022generating,granot2022drop}, weak- or self-supervised learning~\cite{cho2017wssdcnn,tan2019cycle}.
% However, these methods still struggle to preserve both content and structure or generate less visual artifacts (\textit{e.g.,} out-of-boundary, or OOB, artifact) as shown in Fig.~\ref{fig:motivation}. 
% The main issue for these methods is that they apply the \emph{same transformation} to salient and non-salient content.  
 However, these methods struggle to balance content preservation and structure fidelity while minimizing artifacts (\textit{e.g.,} out-of-boundary, or OOB, artifacts; see Fig.~\ref{fig:motivation}). 
 A key limitation is applying the \emph{same transformation} to both salient and non-salient content. 
 Moreover, many recent \emph{test-time optimization} methods~\cite{shaham2019singan,InGAN,hinz2021improved,zhang2022petsgan} suffer from a slow running speed.

To solve these issues, we propose an end-to-end trainable method \methodname (\underline{H}uman-\underline{A}ligned \underline{L}ayered transf\underline{O}rmations) for image retargeting. Unlike previous methods, \methodname warps images using \emph{layered transformations}, recognizing that distortions in salient regions are more perceptible. By decomposing images into salient and non-salient layers via a saliency map, it applies distinct transformations to each, preserving critical details while mitigating OOB issues.
Also, thanks to our end-to-end training strategy, our model enjoys a \emph{significantly faster speed} at the inference time. 

% To further reduce the structure and content loss in output images, we use perceptual loss function as weak supervision to guide the algorithm to produce images close to the original image's content and structure. 
% To produce images close to the input image's content and structure, we study different perceptual loss functions as weak-supervision to guide the algorithm.
% DreamSim~\cite{fu2023dreamsim}, which emphasizes mid-level structure and distortion, is well-suited as a perceptual loss for image retargeting.
% % To overcome the problem of DreamSim
% However, since DreamSim is trained on square images, it cannot be directly applied to image retargeting. 
% To address this, we develop a \emph{layout augmentation} technique 
% that adapts DreamSim for image retargeting
% and we introduce a new loss function, Perceptual Structure Similarity Loss (PSSL), which aligns closely with human perception.
To better preserve content and structure, we explore perceptual loss functions as weak supervision. 
DreamSim~\cite{fu2023dreamsim}, which captures mid-level structure and distortion, is well-suited for image retargeting but is trained \emph{only} on square images, limiting its direct applicability. 
To address this, we introduce \emph{layout augmentation} to adapt DreamSim and propose Perceptual Structure Similarity Loss (PSSL), which better aligns with human perception.

Our contributions are as follows:
\begin{itemize}
    \item A novel end-to-end trainable image retargeting algorithm based on layered transformations achieves significantly faster speed at the inference time.
    \item A new Perceptual Structure Similarity Loss for image retargeting tasks, aligning well with human perception.
    \item Extensive quantitative results and a user study on the RetargetMe dataset demonstrate that \methodname achieves SOTA, in terms of preserving the structure and the content, producing high quality output.
\end{itemize}

\section{Related work}\label{sec:related}
\vspace{-2mm}
\textbf{Image Retargeting.} Image retargeting is a task to generate images with arbitrary aspect-ratios given an input image. 
Over the years, various approaches have been proposed, including conventional optimization-based methods~\cite{rubinstein2008improved,rubinstein2009multi,barnes2009patchmatch,simakov2008summarizing, wolf2007non,pritch2009shift,wang2008optimized,karni2009energy}, weakly-supervised learning~\cite{cho2017weakly,tan2019cycle}, deep reinforcement learning~\cite{kajiura2020self}, GAN based models~\cite{shaham2019singan,InGAN,hinz2021improved,zhang2022petsgan}, Patch Nearest Neighbor (PNN)~\cite{granot2022drop,elnekave2022generating}, and diffusion models~\cite{wang2022sindiffusion,kulikov2023sinddm,zhang2023sine,nikankin2022sinfusion}. 
% Despite these advancements, many of these methods continue to face challenges in preserving the content and structure of the original input image.
% Unfortunately, they still struggle in preserving the content and structure of the input image. 
{
Compared to optimization-based methods, we train an end-to-end model and it has \emph{faster} inference speed. 
Compared to end-to-end methods, our method uses layered transformations and predicts multiple warping flows, avoiding out-of-boundary issues and preserving salient contents better.}
\vspace{-1mm}

\topic{Layered representations.}
Layered representations~\cite{lu2020,lu2021omnimatte,yang2021selfsupervised} enable more flexible manipulation for an image or a video on different layers.
It has been widely used for both images~\cite{he2009single,gandelsman2019double} and videos~\cite{lu2020,Liu-TPAMI-2021,lu2021omnimatte,kasten2021layered,lee2023shape}.
% We adopt the idea of layered representation for an image, and apply different warping fields for a flexible manipulation.
{We adopt the idea of layered representations and use it in the image retargeting task. It avoids out-of-boundary issues in the previous methods.}
\vspace{-1mm}

\topic{Perceptual losses.} 
With the revolution of deep learning, many pretrained networks~\cite{NIPS2012alexnet,simonyan2014very,he2016deep}, can extract meaningful features from the images.
Defined by measuring the feature distances, learning-based metrics~\cite{NIPS2016_371bce7d,johnson2016perceptual,zhang2018perceptual,prashnani2018pieapp} show better alignment with human perception than the classic ones.
% Some perceptual loss functions, e.g., VGG, LPIPS, PIE-APP
% Afterwards, other perceptual loss functions, such as VGG loss~\cite{johnson2016perceptual}, LPIPS~\cite{zhang2018perceptual}, PIE-APP~\cite{prashnani2018pieapp}, show tremendous improvements and align better with human perception.
% Recent DreamSim focusing more on mid-level structures.
More recently, DreamSim~\cite{fu2023dreamsim} is proposed to capture the mid-level similarities, such as structure and layout, between images.
%
% Perceptual losses are also used in image retargeting~\cite{cho2017wssdcnn,tan2019cycle} in the absence of paired training data.
Different from previous retargeting methods~\cite{cho2017wssdcnn,tan2019cycle} which also use perceptual losses,
% We further adapt DreamSim to detect the structural distortions and propose a Perceptual Structure Similarity Loss for image retargeting.
{we use DreamSim, a perceptual loss focusing on the mid-level features such as structures and layouts. We find previous perceptual loss functions (\emph{e.g.}, LPIPS~\cite{zhang2018perceptual}) have difficulties handling structure distortions. We further adapt DreamSim to the image retargeting task by proposing a layout augmentation.}

% \topic{Weakly-supervised learning.}
% Weakly-supervised learning is an approach...The idea is used in many tasks, such as object detection~\cite{bilen2016weakly}, visual relationship learning~\cite{peyre2017weakly}, ...
% For image retargeting, WSSDCNN~\cite{cho2017wssdcnn} first uses perceptual loss as there is no paired training data, followed by other works~\cite{tan2019cycle}.
% Our method also uses a weakly-supervised learning, with additional geometry disturbances. 

\section{Methodology}\label{sec:method}

\subsection{Overview of HALO}
\begin{figure*}[t]
    \begin{center}
    \centering
    \includegraphics[trim=0 0 0 0, clip,width=\textwidth]{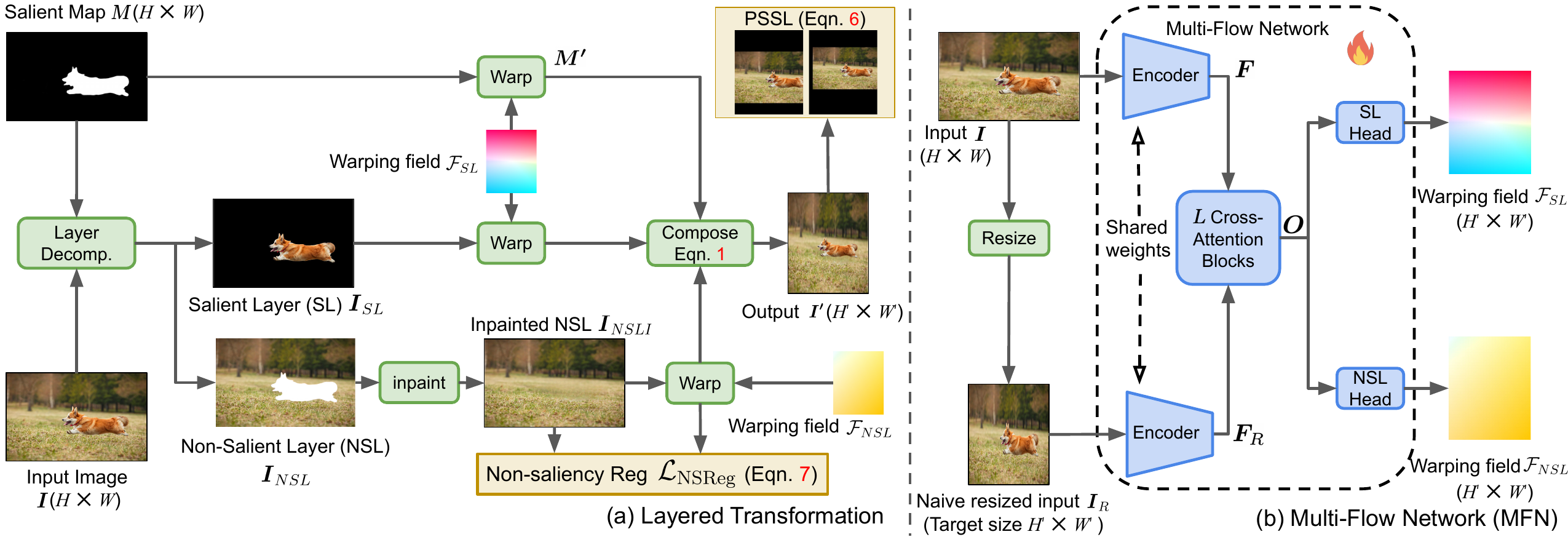}
    \caption{
    \textbf{Overview of HALO.} 
    We retarget an input image $\boldsymbol{I} \in \mathbb{R}^{H \times W}$ to an output image $\boldsymbol{I}^{\prime}$ at the target size $H^{\prime} \times W^{\prime}$.
    (a) \textbf{Layered Transformation.} We decompose the input image into a salient layer (SL) $\boldsymbol{I}_{SL}$ and a non-salient layer (NSL) $\boldsymbol{I}_{NSL}$ with a saliency map from~\cite{gao2024multiscale}.
    We inpaint the hole in $\boldsymbol{I}_{NSL}$ by~\cite{suvorov2022lama} to obtain the inpainted NSL $\boldsymbol{I}_{NSLI}$.
    We then transform $\boldsymbol{I}_{SL}$ and $\boldsymbol{I}_{NSLI}$ with the predicted warping fields $\mathcal{F}_{SL}$ and $\mathcal{F}_{NSL}$, respectively. 
    We also warp the saliency map $\boldsymbol{M}$ with $\mathcal{F}_{SL}$ to obtain a warped saliency map $\boldsymbol{M}^{\prime}$.
    We obtain the output $\boldsymbol{I}^{\prime}$ by composing the warped layers with $\boldsymbol{M}^{\prime}$ via Eqn.~\ref{eqn:layered_output}.
    % We apply \emph{layout disturbance} (Section~\ref{sec:layout_disturb}) to the input image and calculate DreamSim~\cite{fu2023dreamsim} for training.
    To train our model, we use our Perceptual Structure Similarity Loss (PSSL, Eqn.~\ref{eqn:pssl}) and non-saliency regularization (Eqn.~\ref{eqn:NSReg}). 
    (b) \textbf{Multi-Flow Network.} Our Multi-Flow Network (MFN) takes the input image $\boldsymbol{I} \in \mathbb{R}^{H \times W}$ and its resized version $\boldsymbol{I}_{R} \in \mathbb{R}^{H^{\prime} \times W^{\prime}}$ as input.
    $\boldsymbol{I}$ and $\boldsymbol{I}_{R}$ are encoded with a shared encoder.
    The resulting feature maps are then passed into $L$ cross-attention blocks.
    Finally, Salient-Layer (SL) head and Non-Salient Layer (NSL) head predict a salient flow $\mathcal{F}_{SL}$ and a non-salient flow $\mathcal{F}_{NSF}$ for the corresponding layers.
    }
    \label{fig:method_overview}
    \end{center}
    \vspace{-3mm}
\end{figure*}

Fig.~\ref{fig:method_overview} shows the framework of our method. 
\methodname takes an image $\boldsymbol{I} \in \mathbb{R}^{H \times W}$ and its saliency map $\boldsymbol{M}$ as inputs to predict an output image $\boldsymbol{I}^{\prime} \in \mathbb{R}^{H^{\prime} \times W^{\prime}}$,
where $H, W$ are the input height and width, $H^{\prime},W^{\prime}$ are the output height and width. 
The saliency map is a heatmap that measures the importance of pixels in the input image. The saliency map could be generated by a saliency detector~\cite{gao2024multiscale}, a segmentation network (\textit{e.g.,} SAM~\cite{kirillov2023segment}), or a user-defined mask.
In this paper, we follow previous works~\cite{elsner2024retargeting,rubinstein2008improved} and use saliency maps predicted by an off-the-shelf salient detector, MDSAM~\cite{gao2024multiscale}.
However, \methodname also works for other masks (Fig.~\ref{fig:mdsam_fail}).
Given a saliency map $\boldsymbol{M}$, we decompose the input $\boldsymbol{I}$ into a salient layer as 
$\boldsymbol{I}_{SL}$ = $\boldsymbol{I} \odot  \boldsymbol{M}$ and a non-salient layer $\boldsymbol{I}_{NSL}$ = $\boldsymbol{I} \odot (\boldsymbol{1} - \boldsymbol{M}) $, where $\odot$ is the element-wise multiplication. To fill in the holes of the non-salient layer, we inpaint it with an off-the-shelf inpainting model~\cite{suvorov2022lama}: $\boldsymbol{I}_{NSLI}$ = $\text{Inpaint}(\boldsymbol{I}_{NSL})$.

The reason for decomposing an image into two layers is based on the observation that a single transformation, as in~\cite{cho2017wssdcnn,tan2019cycle}, cannot handle both salient and non-salient contents simultaneously well and may result in
out-of-boundary (OOB) issues as shown in Fig.~\ref{fig:motivation} and Fig.~\ref{fig:qual_ablation}. 
A single transformation is able to preserve the salient content to the new target size, but may warp the non-salient pixels in an undesired way.
% Based on this observation, we decompose the input image into two layers based on the saliency map, and ask the model to predict different warping fields.
Applying multiple transformations gives the model more flexibility to achieve retargeting without suffering from the OOB issues (Fig.~\ref{fig:qual_ablation}).
% an image $\boldsymbol{I}$ by two layers,
% salient layer $\boldsymbol{I}_{SL}$ (\textit{e.g.,} obvious objects) and non-salient $\boldsymbol{I}_{NSL}$.
% $\boldsymbol{I}_{SL}$ and $\boldsymbol{I}_{NSL}$, where $\boldsymbol{I}_{SL}$ represents the salient content (\textit{e.g.,} obvious objects) while $\boldsymbol{I}_{NSL}$ represents the non-salient part (\textit{e.g.,} background or duplicated textures). 
% \yl{Seems that we didn't define $\odot$? Maybe it is obvious..}
% \begin{equation} \label{eqn:layered_input}
%     \boldsymbol{I} = \boldsymbol{I}_{SL} \odot \boldsymbol{M} + \boldsymbol{I}_{NSL} \odot(\boldsymbol{1} - \boldsymbol{M}) \,,
%  \end{equation}
% where $\odot$ is the element-wise multiplication.
We finally formulate the output image $\boldsymbol{I}^{\prime}$ as 
\begin{equation} \label{eqn:layered_output}
    \scriptstyle
    \boldsymbol{I}^{\prime} = \mathrm{Warp}(\boldsymbol{I}_{SL}, \mathcal{F}_{SL}) \odot \boldsymbol{M^{\prime}} +   \mathrm{Warp}(\boldsymbol{I}_{NSLI}, \mathcal{F}_{NSL}) \odot (\boldsymbol{1} -  \boldsymbol{M^{\prime}}) \,.
\end{equation}
where $\mathcal{F}_{SL}, \mathcal{F}_{NSL} \in \mathbb{R}^{H^{\prime} \times W^{\prime} \times 2}$ are vector warping fields predicted by our Multi-Flow Network (MFN), and the warped saliency map $\boldsymbol{M^{\prime}} = \mathrm{Warp}(\boldsymbol{M}, \mathcal{F}_{SL})$. 

\subsection{Multi-Flow Network}
Inspired by Spatial Transformer Networks (STNs)~\cite{jaderberg2015spatial,peebles2022gg,ofri2023neural}, we design a Multi-Flow Network (MFN) shown in Fig.~\ref{fig:method_overview} to predict two flow maps.
Our MFN consists of an encoder, $L$ cross-attention blocks, and two heads to predict the warping fields.
To condition our network on the target size (or the aspect-ratio), we first resize the input image $\boldsymbol{I}$ to $\boldsymbol{I}_{R}$ with the target size $H^{\prime} \times W^{\prime}$, and pass both $\boldsymbol{I}$ and $\boldsymbol{I}_{R}$ to the encoder, yielding two feature maps $\boldsymbol{F}, \boldsymbol{F}_{R}$:
\begin{equation}
    \boldsymbol{F} = \text{Encoder}(\boldsymbol{I}), \boldsymbol{F}_{R} = \text{Encoder}(\boldsymbol{I}_{R}) \,.
\end{equation}
We notice the resized input $\boldsymbol{I}_{R}$ already provides the coarse position of each object at the target size, but with a distorted structure.
The input image $\boldsymbol{I}$, however, has undistorted structure but no knowledge about the positions at the target size.
We thus leverage the cross-attention blocks~\cite{croco_v2} to exchange the information between $\boldsymbol{I}$ and $\boldsymbol{I}_{R}$.
Inspired by previous work~\cite{granot2022drop} where each patch in the resized image queries the key patches in the input image via a non-differentiable nearest neighbor search, we adapt this idea and make it differentiable with a cross-attention mechanism.
We consider the resized feature $\boldsymbol{F}_{R}$ as query, the input feature $\boldsymbol{F}$ as both key and value, and apply $L$ cross-attention blocks to them to obtain the output feature map:
\begin{equation}
    \boldsymbol{O} = \underbrace{\text{CrossAttn}_L \circ \cdots \circ \text{CrossAttn}_1 (\boldsymbol{F}_R, \boldsymbol{F})}_{L~\text{blocks}}  \,.
\end{equation}

% Previous work such as Patch NN~\cite{granot2022drop} splits the input (key and value) and the resized input (query) into patches in the pixel space. Patch NN then looks for the nearest neighbor for the query patches by checking the key patches, and finally replaces the query patches with the value patches in the next size level. 
% \yl{I think we are missing one sentence here, e.g. any disadvantage/failure of the patch NN and why we adopt?}
% We adopt this idea by using $L$ cross-attention blocks instead, as the attention mechanism is differentiable.
Finally, two heads predict two vector fields $\mathcal{F}_{SL}, \mathcal{F}_{NSL} \in \mathbb{R}^{H^{\prime} \times W^{\prime} \times 2}$ for warping in Eqn.~\ref{eqn:layered_output}:
\begin{equation} \label{eqn:network}
    % \mathcal{F}_{SL}, \mathcal{F}_{NSL} = \text{MFN}(\boldsymbol{I}, \boldsymbol{I}_{R}) \,,
    \mathcal{F}_{SL} = \text{Head}_{SL}(\boldsymbol{O}),  \mathcal{F}_{NSL} = \text{Head}_{NSL}(\boldsymbol{O}) \,,
\end{equation}
where $\mathcal{F}_{SL}$ is for salient layer and $\mathcal{F}_{NSL}$ for non-salient layer.
Please refer to the \textbf{Supplementary Material} for more details.

% \subsection{Compositional transformations}
% Previous methods~\cite{cho2017wssdcnn,tan2019cycle} uses one single transformation to warp the input images.
% We observe that single transformation cannot handle both salient and non-salient contents well in the image. 
% Sometimes, it brings an out-of-boundary (OOB) issue as shown in Fig.~\ref{fig:qual_ablation}. While the transformation tends to preserve the salient content to the new target size, it also warps the non-salient pixels in an desired way.
% Ideally, we expect a more ``aggressive'' transformation for the salient content, but a ``milder'' transformation for the non-salient part. \yl{this seems a bit subjective and vague to me. Does the aggressive transformation mean accurate, e.g. preserving aspect ratio or with more details?}
% It is hard to achieve this goal with only one transformation.
% Based on this observation, we separate the input image into two layers with the saliency map (Eqn.~\ref{eqn:layered_input}), and ask the network to predict two sampling grids $\mathcal{F}_{SL}, \mathcal{F}_{NSL}$ (Eqn.~\ref{eqn:network}).
% To fill in the holes, we inpaint the background with an off-the-shelf model~\cite{suvorov2022lama}.
% Finally we compose two warped layers with $\mathcal{F}_{SL}, \mathcal{F}_{NSL}$ in Eqn.~\ref{eqn:layered_output}.
% We find this operation gives the network more freedom to achieve a retargeting without suffering from the OOB issues (Fig.~\ref{fig:qual_ablation}).

\subsection{Perceptual Structure Similarity Loss}\label{sec:layout_disturb}
\begin{figure*}[h!]
    \centering
    \begin{subfigure}[t]{0.3\textwidth}
        \centering
        \caption{Source}
        \frame{\includegraphics[trim=0 0 0 0, clip,height=0.115\textheight]{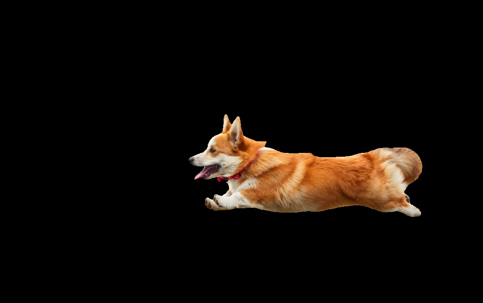}}
        \text{LPIPS sim.$\uparrow$ / DreamSim sim.$\uparrow$}
    \end{subfigure} \hfill
    \begin{subfigure}[t]{0.3\textwidth}
        \centering
        \caption{Translation (no distortion)}
        \frame{\includegraphics[trim=0 0 0 0, clip,height=0.115\textheight]{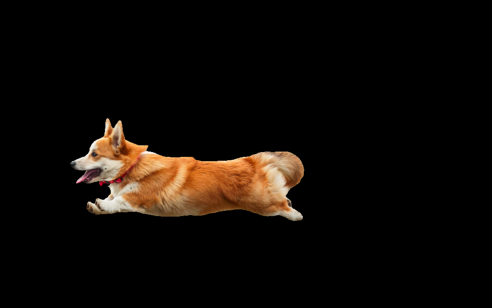}}
        \text{0.8490 / \textbf{0.9578}}
    \end{subfigure} \hfill
    \begin{subfigure}[t]{0.3\textwidth}
        \centering
        \caption{Scale $+$ translation (distorted)}
        \frame{\includegraphics[trim=0 0 0 0, clip,height=0.115\textheight]{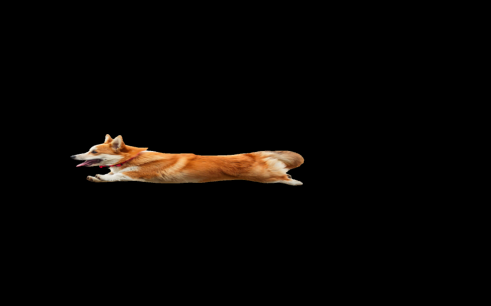}}
        \text{\textbf{0.8722} / 0.8508}
    \end{subfigure}
    
    \caption{\textbf{Comparison between DreamSim and LPIPS.}
    We calculate the similarities of the features from LPIPS~\cite{zhang2018perceptual} and DreamSim~\cite{fu2023dreamsim} for image pairs (a, b) and (a, c), and report the results under each column (LPIPS sim.$\uparrow$ / DreamSim sim.$\uparrow$).
    Surprisingly, the distorted result in (c) shows a higher LPIPS similarity to the source image compared to the undistorted image in (b).
    DreamSim, however, is more sensitive to structural similarity, showing a higher score for the undistorted image pair (a, b) and a lower score for the distorted pair (a, c).
    }
    \label{fig:perceptual_loss_comp}
    \vspace{-3mm}
\end{figure*}

% Lack of paired data -> we need weak supervision -> we need perceptual loss -> we use DreamSim -> Directly use DreamSim still bad -> Introducing layout disturbances.
One of the challenges of training an image retargerting model is the absence of paired data for supervision.
Previous works, such as~\cite{cho2017wssdcnn,tan2019cycle,mastan2020dcil} use a perceptual loss (\textit{e.g.,} VGG loss~\cite{Simonyan14a} or LPIPS~\cite{zhang2018perceptual}) between the input and the output as a weak supervision. 
These perceptual loss functions calculate the distance between feature maps via a pretrained network, and do not enforce a strict supervision as pixelwise $\ell_1$ or $\ell_2$ losses.
However, popular perceptual losses like LPIPS are less sensitive to structural distortions compared to DreamSim~\cite{fu2023dreamsim} in Fig.~\ref{fig:perceptual_loss_comp}.
Therefore, we adopt DreamSim as our perceptual quality metric.
% Since the DreamSim~\cite{fu2023dreamsim} loss focuses more on the mid-level structure, we adopted this loss as the perceptual loss for training.

\begin{figure*}[h!]    
    \centering
    \captionsetup{type=figure}
    \includegraphics[trim=0 0 0 0, clip,width=\textwidth]{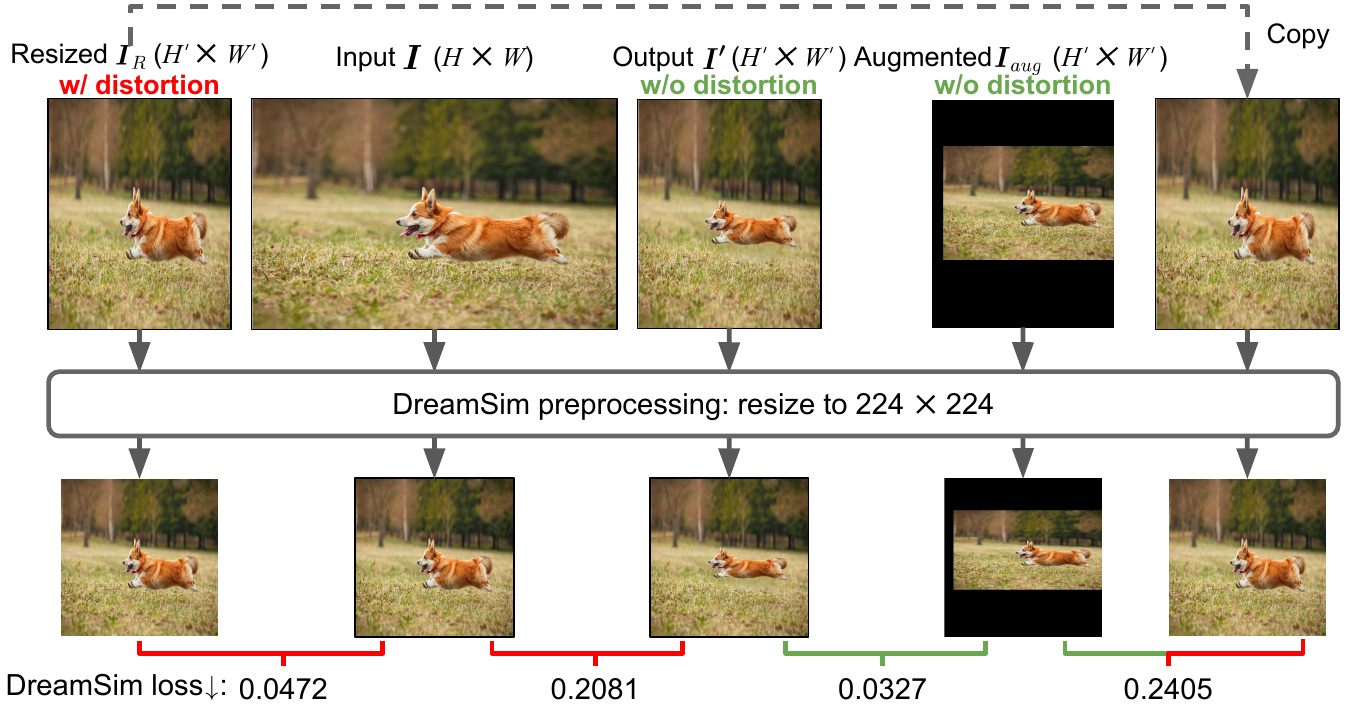}

    \caption{\textbf{Layout Augmentation.} 
    Because DreamSim~\cite{fu2023dreamsim} preprocesses the images by resizing them to $224\times224$, after preprocessing, the naively resized input $\boldsymbol{I}_{R}$ (distorted at the target size $H^{\prime} \times W^{\prime}$) and the input $\boldsymbol{I}$ have a similar structure and result in a small DreamSim loss.
    On the other hand, the layout augmentation $\boldsymbol{I}_{aug}$ (undistorted at the target size) has a small DreamSim loss with the (ideally) undistorted output $\boldsymbol{I}^{\prime}$. 
    % The (ideally) undistorted output $\boldsymbol{I}^{\prime}$ has a large DreamSim loss with the input $\boldsymbol{I}$.
    Therefore, to obtain an undistorted output, we compute the DreamSim loss between the output $\boldsymbol{I}^{\prime}$ and $\boldsymbol{I}_{aug}$ as supervision, instead of between $\boldsymbol{I}^{\prime}$ and $\boldsymbol{I}$.
    % Directly using the DreamSim loss between the output $\boldsymbol{I}^{\prime}$ and $\boldsymbol{I}$ as supervision leads to a distorted result similar to $\boldsymbol{I}_{R}$.
    % In contrast, $\boldsymbol{I}_{aug}$ has no distortion at the target size.
    % Therefore, we compute the DreamSim loss between the output $\boldsymbol{I}^{\prime}$ and $\boldsymbol{I}_{aug}$ instead.
    }
    \label{fig:why_disturb}
    \vspace{-3mm}
\end{figure*}

% Directly using DreamSim for training still introduces distortion,
% (See Fig.~\ref{fig:qual_ablation}, w/o layout disturb.), 
Unfortunately, directly using DreamSim does not work for image retargeting, since DreamSim is trained on square, undistorted images and preprocesses images by resizing them to a fixed square size $224 \times 224$. 
As shown in Fig.~\ref{fig:why_disturb}, 
% the resized $\boldsymbol{I}_{R}$ (at $224 \times 224$) exhibits structural distortion compared with input image $\boldsymbol{I}$, but due to DreamSim's preprocessing step, the DreamSim loss is very small.
the preprocessed $\boldsymbol{I}_{R}$ (at $224 \times 224$) exhibits a very small DreamSim loss with the input image $\boldsymbol{I}$, despite $\boldsymbol{I}_{R}$ having distortion at the target size.
Consequently,
supervising the training with DreamSim loss between the input $\boldsymbol{I}$ and the output leads to a similar, distorted output as $\boldsymbol{I}_{R}$ at the target size.
This makes the original DreamSim loss not suitable for image retargeting.

% This suppresses the distortion-awareness of DreamSim as shown in Fig.~\ref{fig:why_disturb}.

To adapt DreamSim to image retargeting, we propose to apply a random layout transformation (with {scaling ${s}$ and translation $\boldsymbol{t}$}) to disturb the input $\boldsymbol{I}$ at the target size $H^{\prime} \times W^{\prime}$ as an augmentation.
\begin{equation}\label{eqn:layout_disturb}
    \boldsymbol{I}_{aug} = \mathrm{Warp}(\boldsymbol{I}, \mathcal{F}(s, \boldsymbol{t})) \,,
\end{equation}
where $\boldsymbol{I}_{aug} \in {H^{\prime} \times W^{\prime}}$, and the warping field $\mathcal{F}(s, t) \in \mathbb{R}^{H^{\prime} \times W^{\prime} \times 2}$ is determined by the scaling factor ${s}$ and a 2D translation $\boldsymbol{t}=[t_1, t_2]$ both drawn from uniform distributions.
This results in images $\boldsymbol{I}_{aug}$ \emph{without distortions} at target size $H^{\prime} \times W^{\prime}$ as shown in Fig.~\ref{fig:method_overview} and Fig.~\ref{fig:why_disturb}.
We encourage readers to refer to the \textbf{Supplementary Material} for more examples from the layout augmentation.
We use $\boldsymbol{I}_{aug}$ as a pseudo ground truth and leverage DreamSim's structure-awareness as supervision during training and denote \textbf{Perceptual Structure Similarity Loss} (PSSL) as
\begin{equation}\label{eqn:pssl}
    \mathcal{L}_{PSSL}(\boldsymbol{I}^{\prime}, \boldsymbol{I}) = \mathcal{L}_{\mathrm{DreamSim}}(\boldsymbol{I}^{\prime}, \boldsymbol{I}_{aug}) \,.
\end{equation}

\subsection{Training loss}
\topic{PSSL.} As described in Section~\ref{sec:layout_disturb}, we use PSSL as our main training loss.
We also study the popular LPIPS~\cite{zhang2018perceptual} and demonstrate that DreamSim~\cite{fu2023dreamsim} works better than the LPIPS loss for the image retargeting (See Fig.~\ref{fig:qual_ablation} and Tab.~\ref{tab:ablation}).

\topic{Non-saliency regularization.} We further observe that, although the layered transformations (Eqn.~\ref{eqn:layered_output}) significantly mitigate the OOB issue, some extreme cases still yield OOB artifacts (See Fig.~\ref{fig:qual_ablation}, w/o $\mathcal{L}_{\mathrm{NSReg}}$). 
The OOB issue primarily comes from the inpainted non-salient layer $\boldsymbol{I}_{NSLI}$.
We use a pixelwise $\ell_2$ loss between the warped inpainted non-salient layer and the original one to encourage a mild transformation:
\begin{equation}\label{eqn:NSReg}
    \scriptstyle
    \mathcal{L}_{\mathrm{NSReg}} = \frac{1}{N_{\mathrm{pixel}}} ||\boldsymbol{I}_{NSLI} - \mathrm{Resize}(\mathrm{Warp}(\boldsymbol{I}_{NSLI}, \mathcal{F}_{NSL})) ||_2 \,,
\end{equation}
where we resize the warped inpainted non-salient layer to the same size of $\boldsymbol{I}_{NSLI}$, and $N_{\mathrm{pixel}}$ is the total number of pixels in $\boldsymbol{I}_{NSLI}$.

\topic{Total loss.}
Our training loss is
\begin{equation}\label{eqn:total_loss}
    \mathcal{L}_{\mathrm{total}} = \mathcal{L}_{PSSL} + \lambda_{\mathrm{NSReg}} \mathcal{L}_{\mathrm{NSReg}} \,,
\end{equation}
where PSSL serves as our main loss,  $\mathcal{L}_{\mathrm{NSReg}}$ is a non-saliency regularization regularization term, and $\lambda_{\mathrm{NSReg}}$ controls the strength of $\mathcal{L}_{\mathrm{NSReg}}$. In practice, we use $\lambda_{\mathrm{NSReg}}=2.0$.

\section{Experimental results}\label{sec:results}

\subsection{Setup}
\topic{Dataset.} We train our model on the UHRSD dataset~\cite{xie2022UHRSD}, which consists of $4,932$ training images and $988$ test images. 
Each image comes with an annotated saliency map.
It covers diverse image categories including natural landscapes, street views, and animals. 
%We train our network on the training set of UHRSD.
During training, we resize the images so that their shorter side is scaled to $512$. For example, if the height is greater than the width, the image is rescaled to $(512 \times \frac{\text{height}}{\text{width}}, 512)$.
% We further group the images based on their aspect ratios, allowing us to train the network with batches where images share the same aspect ratio.
We group the images by their aspect ratios and sample images from the same group into each batch.
We test our model and compare with other baseline approaches on the common RetargetMe~\cite{rubinstein2010retargetme} benchmark. 
RetargetMe contains 80 images with different scaling factors ($0.50, 0.75$ and $1.25$) for the test.

\begin{figure*}[t]    
    \centering
    \captionsetup{type=figure}
    % \mpage{0.02}{\raisebox{50pt}{\rotatebox{90}{Input}}}
     \begin{minipage}[t]{0.185\textwidth}
        \centering
        \frame{\includegraphics[trim=0 0 0 0, clip, width=\textwidth]{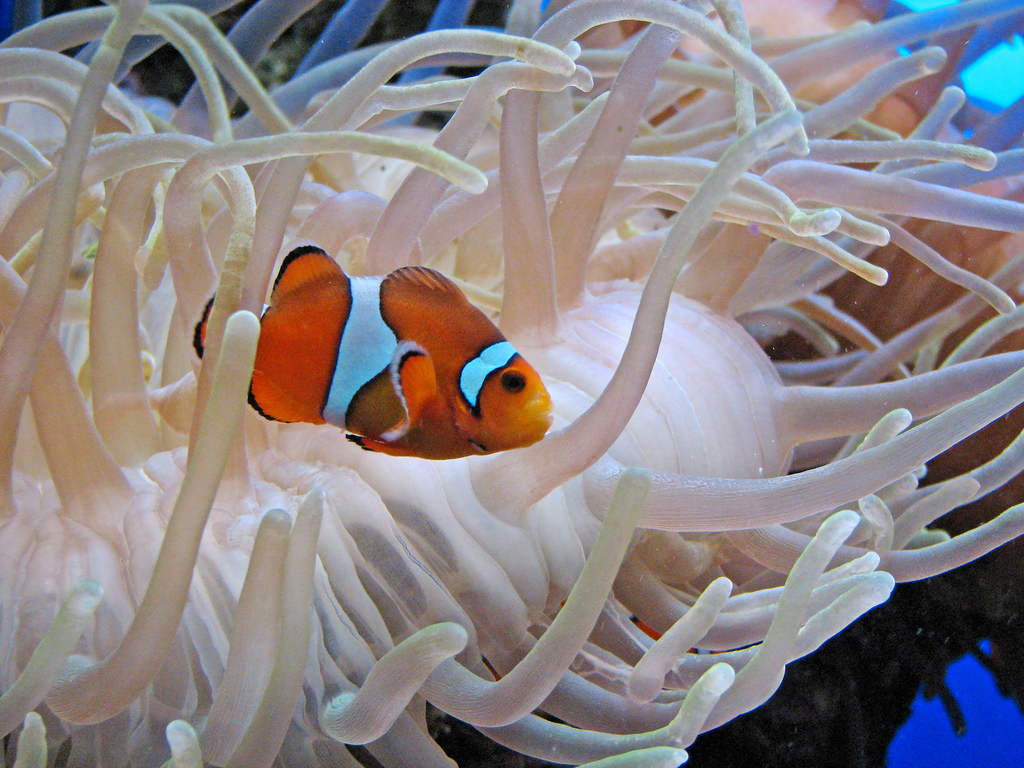}}\\
        \frame{\includegraphics[trim=0 0 0 0, clip, width=\textwidth]{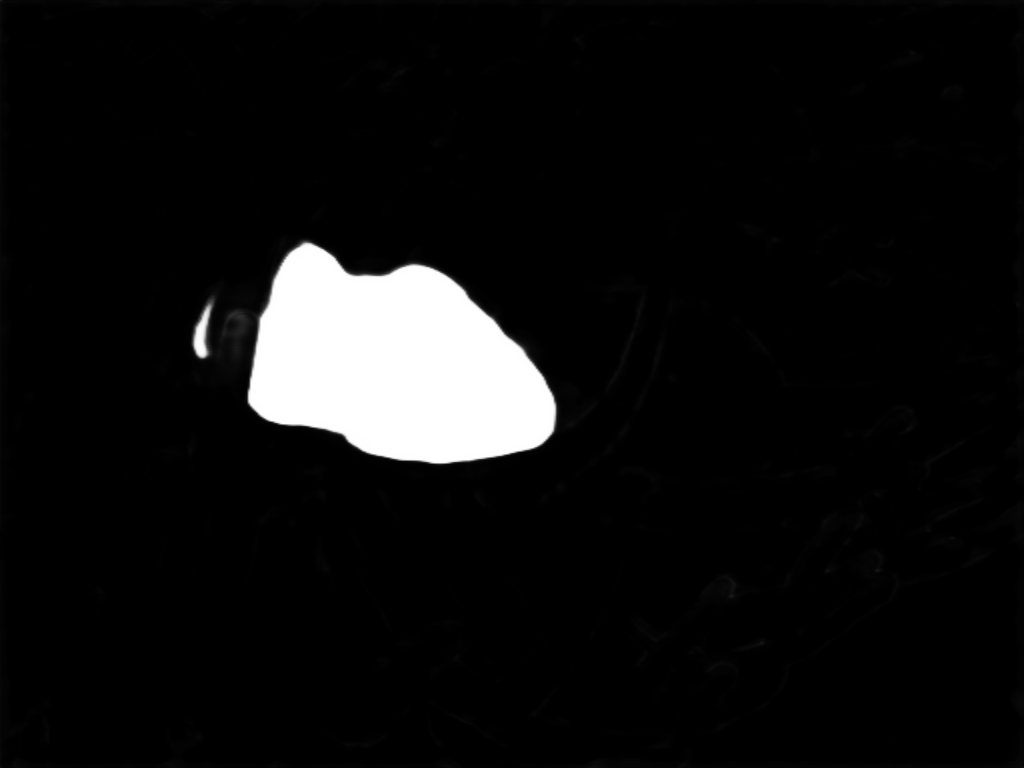}}
    \end{minipage}\hfill
    \begin{minipage}[t]{0.185\textwidth}
        \centering
         \vspace{-24mm}
        \frame{\includegraphics[trim=0 0 0 0, clip,width=\textwidth]{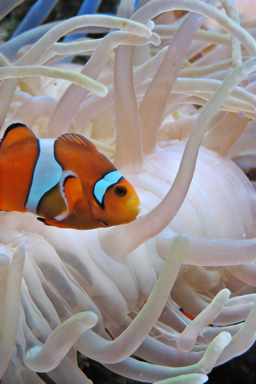}}
    \end{minipage}\hfill
    \begin{minipage}[t]{0.185\textwidth}
        \centering
         \vspace{-24mm}
        \frame{\includegraphics[trim=0 0 0 0, clip,width=\textwidth]{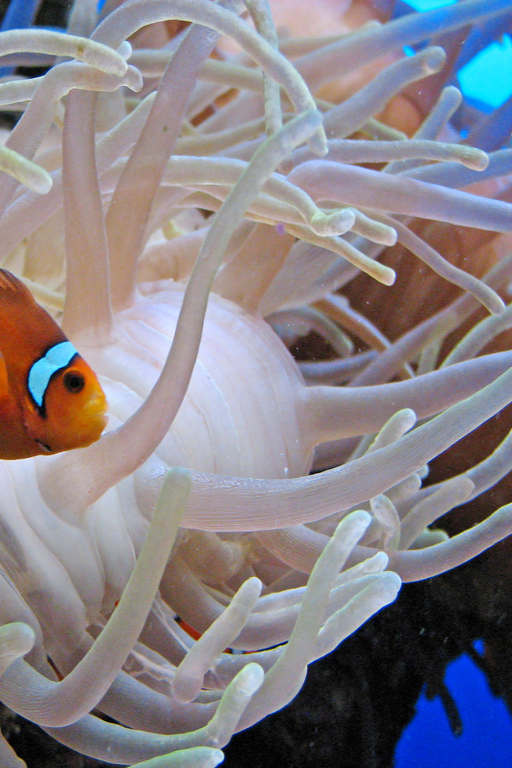}}
    \end{minipage}\hfill
    % \begin{minipage}[t]{0.185\textwidth}
    %     \centering
    %      \vspace{-24mm}
    %     \frame{\includegraphics[trim=0 0 0 0, clip,width=\textwidth]{images/qualitative_comp_assets/fish_dragon.png}}
    % \end{minipage}\hfill
    \begin{minipage}[t]{0.185\textwidth}
        \centering
         \vspace{-24mm}
        \frame{\includegraphics[trim=0 0 0 0, clip,width=\textwidth]{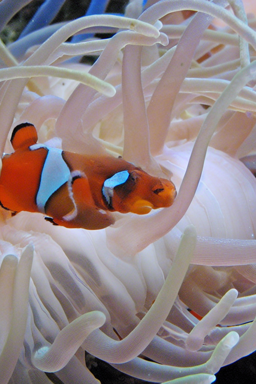}}
    \end{minipage}\hfill
    \begin{minipage}[t]{0.185\textwidth}
        \centering
         \vspace{-24mm}
        \frame{\includegraphics[trim=0 0 0 0, clip,width=\textwidth]{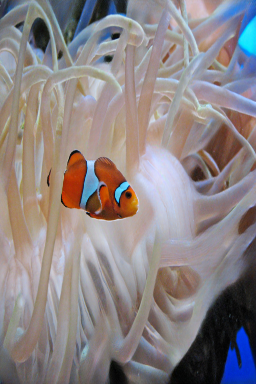}}
    \end{minipage} 
    \vspace{1mm} % Adjust the vertical space as needed
    
     \begin{minipage}[t]{0.185\textwidth}
        \centering
        \frame{\includegraphics[trim=0 0 0 0, clip, width=\textwidth]{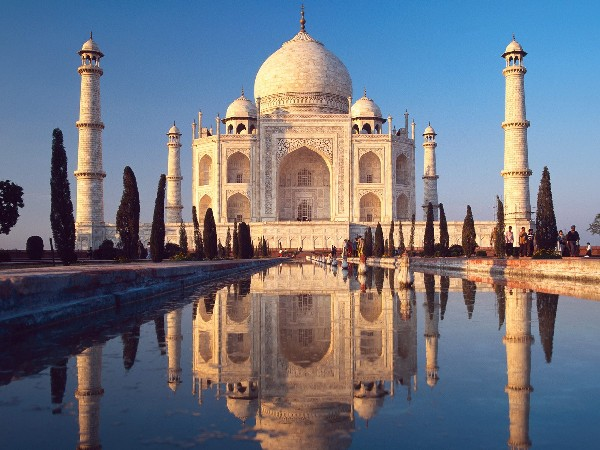}}
        \frame{\includegraphics[trim=0 0 0 0, clip, width=\textwidth]{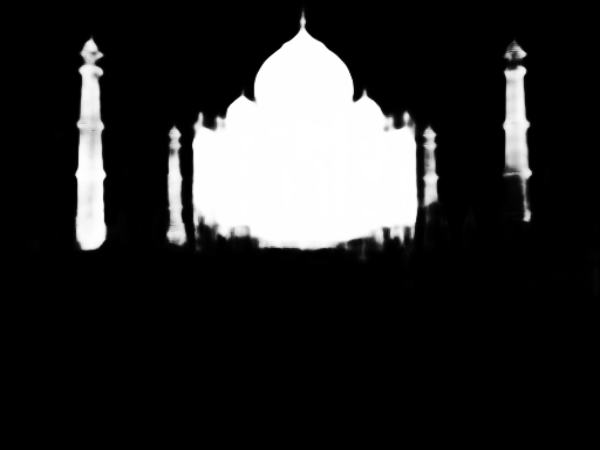}}
        \text{{Input}}
    \end{minipage}  \hfill
    \begin{minipage}[t]{0.185\textwidth}
        \centering
        \vspace{-24mm}
        \frame{\includegraphics[trim=0 0 0 0, clip, width=\textwidth]{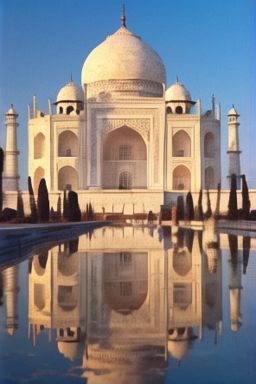}}
        \text{{GPNN~\cite{granot2022drop}}}
    \end{minipage}\hfill
    \begin{minipage}[t]{0.185\textwidth}
        \centering
        \vspace{-24mm}
        \frame{\includegraphics[trim=0 0 0 0, clip, width=\textwidth]{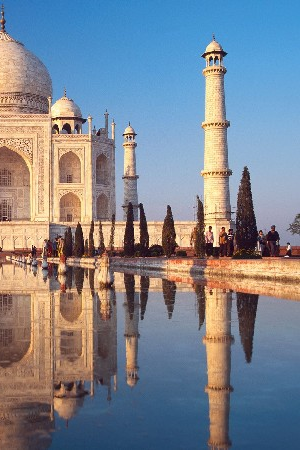}}
        \text{{Self-Play-RL~\cite{kajiura2020self}}}
        % \caption{}
    \end{minipage}\hfill
    % \begin{minipage}[t]{0.185\textwidth}
    %     \centering
    %     \vspace{-24mm}
    %     \frame{\includegraphics[trim=0 0 0 0, clip, width=\textwidth]{images/qualitative_comp_assets/tajmahal_0.50_GPDM.png}}
    %     \text{GPDM~\cite{elnekave2022generating}}
    %     % \caption{{Input frame \#1}}
    % \end{minipage}\hfill
    \begin{minipage}[t]{0.185\textwidth}
        \centering
        \vspace{-24mm}
        \frame{\includegraphics[trim=0 0 0 0, clip, width=\textwidth]{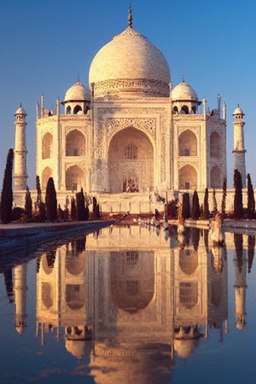}}
        \text{{DragonDiffusion~\cite{mou2024dragondiffusion}}}
        % \caption{{Input frame \#1}}
    \end{minipage}\hfill
    \begin{minipage}[t]{0.185\textwidth}
        \centering
        \vspace{-24mm}
        \frame{\includegraphics[trim=0 0 0 0, clip, width=\textwidth]{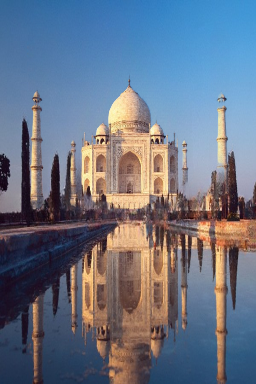}}
        % \caption{}
        \text{Ours}
    \end{minipage} 
    \vspace{1mm} % Adjust the vertical space as needed

    \caption{\textbf{Qualitative comparison.} 
    We compare our method with state-of-the-art image retargeting methods: GPNN~\cite{granot2022drop}, Self-Play-RL~\cite{kajiura2020self}, DragonDiffusion~\cite{mou2024dragondiffusion}.
    We show the input image and its saliency map from~\cite{gao2024multiscale} in the first column.
    Our model preserves the structure and the content of the input images. 
    Notably, in the ``fish'' case, our model is aware of the \emph{affordance} between the fish and the sea anemone.  
    }
    
    \label{fig:qualitative_comp}
    \vspace{-3mm}
\end{figure*}

\topic{Evaluation metrics.}
Previous evaluations~\cite{cho2017wssdcnn,tan2019cycle,kajiura2020self} on image retargeting have relied heavily on user studies. 
Given the rapid advancements of the recent visual representation learning, we propose to use pretrained networks to predict the image features and assess the quality of the outputs based on these features.
We use CLIP image embeddings~\cite{radford2021clip} for the \textbf{content} similarity evaluation. 
We compute the similarity between the input image embedding and the output image embedding.  
To assess \textbf{structure} consistency, we use DreamSim similarity~\cite{fu2023dreamsim}, which focuses on mid-level differences such as structure and layout.
We use the original DreamSim since we do not wish to introduce randomness from~Eqn.~\ref{eqn:layout_disturb} into evaluations.
% We use the VILA score~\cite{ke2023vila} for \textbf{aesthetics} evaluation.
We use VILA~\cite{ke2023vila} and MUSIQ score~\cite{ke2021musiq} for \textbf{aesthetics} evaluation. To better align with other metrics, such as DreamSim and CLIP similarity, we re-normalize VILA and MUSIQ score as a percentage.
Image retargeting also requires to minimize \textbf{visual artifacts}, such as object distortion, missing or duplicated contents, or OOB artifacts. 
Since current assessment models struggle to reliably detect these artifacts, we conduct a user study (Tab.~\ref{tab:user_study_artifacts}) where participants select the output with the best image quality.
% It is still challenging for the current assessment models to detect these artifacts. Therefore, we conduct a user study by asking users to select the output with the best image quality. 
% \yl{best image quality can be aesthetics or visual pleasing or artifacts. What is actual question asked? We may need to be careful here.}
% \yiran{Direct readers to the supp.}
% We use user preferences across different methods as a metric for visual artifact evaluation. We include details about our user study in the {Supplementary Material}.

\subsection{Implementation details}
\topic{Model.} For the encoder of our MFN, we adopt the same CNN-based encoder as in~\cite{peebles2022gg}. We then use $L=3$ cross-attention blocks. For each output head, we predict an affine transformation matrix and convert it into a sampling grid. Please refer to our \textbf{Supplementary Material} for details.

% Hyperparameters.
% lr, batchsize, training sampler, iters...
% GPUs, training time.
\topic{Hyperparameters.} We train our network with an initial learning rate $\alpha=1 \times 10^{-4}$ and an Adam optimizer~\cite{adam}. The learning rate decays by a factor of $0.9$ in every 1000 iterations. 
We use a batch size of 32 and train the model for 200 epochs.
During the training, we sample a random target factor from $\{0.50, 0.75, 1.25, 1.50\}$ for each batch. 
We then randomly choose to change the height or the width of the image with the sampled factor for the current batch.
For example, if we choose to change width and pick a factor $0.50$, we aim to change images' width to its half in this batch.
We train our model with 2 NVIDIA A100 40GB GPUs for around 2 days.

\subsection{Comparison with previous methods}
We compare with three different lines of works: 
\begin{itemize}
    \item Optimization, including SINE~\cite{zhang2023sine}, SinDDM~\cite{kulikov2023sinddm}, GPDM~\cite{elnekave2022generating}, GPNN~\cite{granot2022drop} and IDF~\cite{elsner2024retargeting};
    \item Feed-forward approaches including Self-Play-RL~\cite{kajiura2020self}, Cycle-IR~\cite{tan2019cycle}, WSSDCNN~\cite{cho2017wssdcnn} and PR~\cite{shen2024prune}. 
    \item Editing models. We compare with two drag editing methods, MagicFixup~\cite{alzayer2024magic} and DragonDiffusion~\cite{mou2024dragondiffusion}.
    Please refer to our supplementary material for details.
    % We first place the input at the center of a black canvas with the target size, and then outpaint the boundary with LAMA~\cite{suvorov2022lama} if necessary. Finally we use a drag editing method to adjust the scale and the location of the salient objects with a mask from the saliency detector~\cite{gao2024multiscale}. The scaling factor is calculated by $\frac{H^\prime W^\prime}{HW}$, and the translation by the shift of the centroid of the saliency mask.
\end{itemize}

\begin{table*}[h!]
  \centering
  \scriptsize	
  \caption{\textbf{Quantitative comparison.}
      We compare our method with different types of methods, including optimization-based methods, feed-forward prediction, and drag-style editing, on the RetargetMe dataset~\cite{rubinstein2010retargetme}.
      The test-time runtime for each method is measured on a $1024 \times 813$ image using a single NVIDIA A100 GPU.
      We compute the CLIP~\cite{radford2021clip} embedding similarity to measure content similarity, DreamSim~\cite{fu2023dreamsim} to measure structure similarity, and MUSIQ~\cite{ke2021musiq} and VILA~\cite{ke2023vila} to measure aesthetics.
      To compute the average scores, we normalize MUSIQ and VILA to percentages. 
      % For NIQE, we use $100 - \mathrm{norm(NIQE)}$, as a lower NIQE score indicates better performance.
  }
  % \begin{tabular}{l|c|c|c|cc|c}
  %   \toprule
  %   & \multicolumn{1}{c}{ } & \multicolumn{1}{|c|}{Content} & \multicolumn{1}{c|}{Structure}  & \multicolumn{2}{c|}{Aesthetics}   & \\
  %   \midrule
  %          & Runtime(s.) $\downarrow$ & CLIP sim.(\%)$\uparrow$ & DreamSim sim.(\%)$\uparrow$ & MUSIQ(\%)$\uparrow$ & VILA(\%)$\uparrow$ & Average(\%)$\uparrow$ \\
  %   \midrule
  %     SINE~\cite{zhang2023sine} & 4550.0 & 53.3 & 59.6 & 49.2 & 44.5  & 51.7 \\
  %   SinDDM~\cite{kulikov2023sinddm} & 17424.0 & 79.1 & 40.1 & 36.0 & 24.8  & 45.0 \\
  %   GPDM~\cite{elnekave2022generating} & 61.7 & 53.6 & 65.5 & 48.5 & 47.3  & 53.7 \\
  %   GPNN~\cite{granot2022drop} & 21.3 & 88.5 & \underline{77.5} & 50.7 & \textbf{50.7} &  \underline{66.8} \\
  %   IDF~\cite{elsner2024retargeting} & $>$5400 & \\ 
  %   \midrule
  %   Self-Play-RL~\cite{kajiura2020self} & 1.30 & 88.7 & 76.2 & \textbf{52.1} & \textbf{50.7} & \underline{66.8} \\
  %   Cycle-IR~\cite{tan2019cycle} & 1.01 & 86.7 & 77.0 & 50.4 & 45.2 & 64.8 \\
  %   WSSDCNN~\cite{cho2017wssdcnn} & \underline{0.79} & 85.4 & 69.6 & 41.8 & 33.0 & 57.5 \\
  %   PR~\cite{shen2024prune}  & 10.9 & \textbf{92.3} & 71.9 & 48.5 & 44.4  & 64.3\\
  %   \midrule
  %   MagicFixup~\cite{alzayer2024magic} & 11.0 & 84.8 & 70.1 & 47.1 & 42.4 & 61.6 \\
  %   DragonDiffusion~\cite{mou2024dragondiffusion} & 17.5 & {89.4} & 66.8 & 51.1 & 47.1 & 63.6 \\
  %   \midrule
  %   \methodname (Ours) & \textbf{0.59} & \underline{90.2} & \textbf{78.0} & \underline{51.5} & \underline{48.1} & \textbf{67.0} \\
  %   \bottomrule
  % \end{tabular}%
  \begin{tabular}{c|lc|c|c|cc|c}
    \toprule
      & \multicolumn{2}{c}{ } & \multicolumn{1}{|c|}{Content} & \multicolumn{1}{c|}{Structure}  & \multicolumn{2}{c|}{Aesthetics}   & \\
    \midrule
    \rotatebox{90}{} & Method & Runtime(s.) $\downarrow$ & CLIP sim.(\%)$\uparrow$ & DreamSim sim.(\%)$\uparrow$ & MUSIQ(\%)$\uparrow$ & VILA(\%)$\uparrow$ & Average(\%)$\uparrow$ \\
    \midrule
    \multirow{5}{*}{\rotatebox{90}{\textbf{Optimization}}} 
      & SINE~\cite{zhang2023sine} & 4550.0 & 53.3 & 59.6 & 49.2 & 44.5  & 51.7 \\
    & SinDDM~\cite{kulikov2023sinddm} & 17424.0 & 79.1 & 40.1 & 36.0 & 24.8  & 45.0 \\
    & GPDM~\cite{elnekave2022generating} & 61.7 & 53.6 & 65.5 & 48.5 & 47.3  & 53.7 \\
    & GPNN~\cite{granot2022drop} & 21.3 & 88.5 & \underline{77.5} & 50.7 & \textbf{50.7} &  \underline{66.8} \\
    & IDF~\cite{elsner2024retargeting} & $>$5400 & \textbf{94.2} & 74.2 & 48.8 & 43.6 & 65.2 \\ 
    \midrule
    \multirow{4}{*}{\rotatebox{90}{\textbf{Feed-fwd.}}} 
    & Self-Play-RL~\cite{kajiura2020self} & 1.30 & 88.7 & 76.2 & \textbf{52.1} & \textbf{50.7} & \underline{66.8} \\
    & Cycle-IR~\cite{tan2019cycle} & 1.01 & 86.7 & 77.0 & 50.4 & 45.2 & 64.8 \\
    & WSSDCNN~\cite{cho2017wssdcnn} & \underline{0.79} & 85.4 & 69.6 & 41.8 & 33.0 & 57.5 \\
    & PR~\cite{shen2024prune}  & 10.9 & \underline{92.3} & 71.9 & 48.5 & 44.4  & 64.3\\
    \midrule
    \multirow{2}{*}{\rotatebox{90}{\textbf{Edit.}}} 
    & MagicFixup~\cite{alzayer2024magic} & 11.0 & 84.8 & 70.1 & 47.1 & 42.4 & 61.6 \\
    & DragonDiffusion~\cite{mou2024dragondiffusion} & 17.5 & {89.4} & 66.8 & 51.1 & 47.1 & 63.6 \\
    \midrule
    \rotatebox{90}{\textbf{}} & \methodname (Ours) & \textbf{0.59} & {90.2} & \textbf{78.0} & \underline{51.5} & \underline{48.1} & \textbf{67.0} \\
    \bottomrule
  \end{tabular}%
  \label{tab:quant_full_supp}%
\end{table*}%

\begin{table}[h!]
    \centering
    \small
    \caption{\textbf{User study.}
    Our method HALO is preferred by users by a large margin.
    }
    \begin{tabular}{lcc}
        \toprule
        Comparison & Ours (\%)$\uparrow$ & Baseline (\%)$\uparrow$ \\ 
        \midrule
        Ours vs. Self-Play-RL~\cite{kajiura2020self} & \textbf{56.3} & 43.7 \\ 
        Ours vs. GPNN~\cite{granot2022drop} & \textbf{53.8} & 46.2 \\ 
        Ours vs. DragonDiffusion~\cite{mou2024dragondiffusion} & \textbf{67.4} & 32.6 \\ 
        \midrule
        Average  &  \textbf{59.2} &  40.8 \\
        \bottomrule
    \end{tabular}
    \label{tab:user_study_artifacts}
    \vspace{-3mm}
\end{table}

% \begin{tabular}{cc}
    % \toprule
    %       & User preference ($\%$)  \\
    % \midrule
    % GPDM~\cite{elnekave2022generating}     &  5.47 \\
    % Self-Play-RL~\cite{kajiura2020self}     &  20.86 \\
    % MagicFixup~\cite{alzayer2024magic} & 30.23  \\
    % \methodname (Ours)           &   \textbf{43.44}   \\
    % \bottomrule
    % \end{tabular}%

\topic{Quantitative evaluation.} We report quantitative evaluation results in Tab.~\ref{tab:quant_full_supp}.
Our method achieves the better performance in terms of content and structure preservation.
% While it performs slightly worse than Self-Play-RL on aesthetics, 
Our model outperforms all others when averaging across all four metrics, yielding the highest overall score.
% our model is slightly worse than GPNN and Self-Play-RL, because the criterion model VILA~\cite{ke2023vila} has not seen distorted images during its training. 
% \yl{Any chance we can win over using Musiq?}
% \yiran{Tried MUSIQ where we are still at the second place}
Notably, compared to optimization-based generative models, our approach enjoys faster inference speed while achieving superior performance.

\topic{User study.}
% We also conduct a user study among 16 participants on all 80 images (1280 votes) in RetargetMe~\cite{rubinstein2010retargetme}.
We further conduct a user study on all 80 images from the RetargetMe dataset~\cite{rubinstein2010retargetme} against three methods, Self-Play-RL~\cite{kajiura2020self}, GPNN~\cite{granot2022drop} and DragonDiffusion~\cite{mou2024dragondiffusion}.
For each category in Tab.~\ref{tab:quant_full_supp}, we compare our method with the best-performing baseline.
In each comparison, users are asked to choose between our method and one of the baselines. Each image receives 5 votes, resulting in a total of 1,200 votes. The results are summarized in Tab.~\ref{tab:user_study_artifacts}.
We report the results in Tab.~\ref{tab:user_study_artifacts}.
% We exclude the recent methods PR~\cite{shen2024prune} and IDF~\cite{elsner2024retargeting} from the study, as they perform noticeably worse in Tab.~\ref{tab:quant_full_supp}.
Our model receives significantly higher user preference compared to other baselines, 
suggesting that it better aligns with human perception.

\begin{figure*}[h!]    
    \centering
    \captionsetup{type=figure}
    % \mpage{0.02}{\raisebox{50pt}{\rotatebox{90}{Input}}}
    \begin{minipage}[t]{0.16\textwidth}
        \centering
         {\includegraphics[trim=0 0 0 0, clip,width=\textwidth]{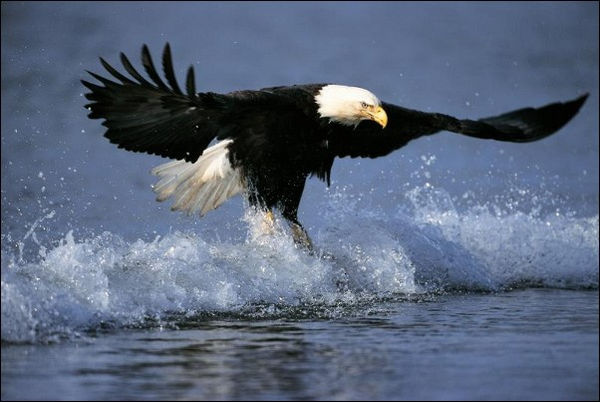}}
    \end{minipage}\hfill
    \begin{minipage}[t]{0.16\textwidth}
        \centering
        {\includegraphics[trim=0 0 0 0, clip,width=\textwidth]{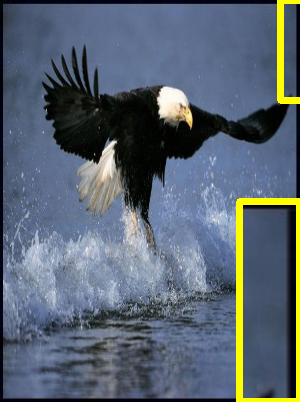}}
    \end{minipage}\hfill
    \begin{minipage}[t]{0.16\textwidth}
        \centering
         {\includegraphics[trim=0 0 0 0, clip,width=\textwidth]{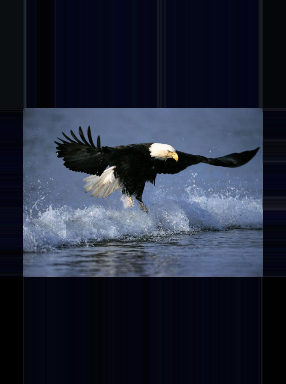}}
        % \caption{}
    \end{minipage}\hfill
    \begin{minipage}[t]{0.16\textwidth}
        \centering
         {\includegraphics[trim=0 0 0 0, clip,width=\textwidth]{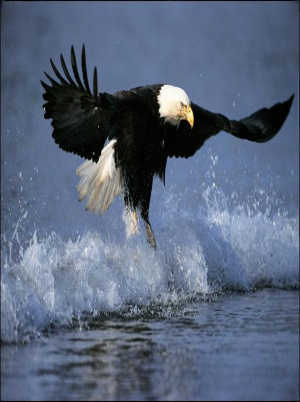}}
        % \caption{}
    \end{minipage}\hfill
    \begin{minipage}[t]{0.16\textwidth}
        \centering
         {\includegraphics[trim=0 0 0 0, clip,width=\textwidth]{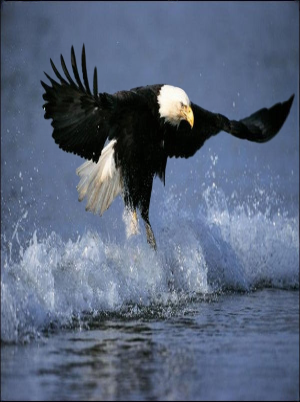}}
        % \caption{}
    \end{minipage}\hfill
    \begin{minipage}[t]{0.16\textwidth}
        \centering
         {\includegraphics[trim=0 0 0 0, clip,width=\textwidth]{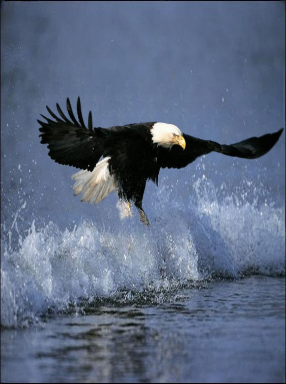}}
        % \caption{}
    \end{minipage}

    % Break to the next row
    % \vspace{1mm} % Adjust the vertical space as needed
    % % \mpage{0.02}{\raisebox{50pt}{\rotatebox{90}{Output}}}
    % \begin{minipage}[t]{0.16\textwidth}
    %     \centering
    %      {\includegraphics[trim=0 0 0 0, clip,width=\textwidth]{images/ablation_assets/Canalhouse.png}}
    %     % \caption{{Input frame \#1}}
    % \end{minipage}\hfill
    % \begin{minipage}[t]{0.16\textwidth}
    %     \centering
    %      {\includegraphics[trim=0 0 0 0, clip,width=\textwidth]{images/ablation_assets/Canalhouse_0.50_singleTrans_zoom.png}}
    %     % \caption{{Input frame \#2}}
    % \end{minipage}\hfill
    % \begin{minipage}[t]{0.16\textwidth}
    %     \centering
    %      {\includegraphics[trim=0 0 0 0, clip,width=\textwidth]{images/ablation_assets/Canalhouse_0.50_noRegBg.png}}
    %     % \caption{Output frame \#2}
    % \end{minipage}\hfill
    % \begin{minipage}[t]{0.16\textwidth}
    %     \centering
    %      {\includegraphics[trim=0 0 0 0, clip,width=\textwidth]{images/ablation_assets/Canalhouse_0.50_noDisturb.png}}
    %     % \caption{}
    % \end{minipage}\hfill
    % \begin{minipage}[t]{0.16\textwidth}
    %     \centering
    %      {\includegraphics[trim=0 0 0 0, clip,width=\textwidth]{images/ablation_assets/Canalhouse_0.50_LPIPS.png}}
    %     % \caption{}
    % \end{minipage}\hfill
    % \begin{minipage}[t]{0.16\textwidth}
    %     \centering
    %      {\includegraphics[trim=0 0 0 0, clip,width=\textwidth]{images/ablation_assets/Canalhouse_0.50_ours.png}}
    %     % \caption{}
    % \end{minipage}

    \mpage{0.15}{{Input}}\hfill
    \mpage{0.15}{{Single Transformation}}\hfill
    \mpage{0.15}{{w/o $\mathcal{L}_{\mathrm{NSReg}}$}}\hfill
    \mpage{0.15}{{w/o augmentation}}\hfill
    \mpage{0.15}{{Ours (w/ LPIPS)}}\hfill
    \mpage{0.15}{{Ours (w/ DreamSim)}}
    % \vspace{-15mm}
    
    \caption{\textbf{Ablation study.} 
    We show the effect of each component by removing one component each time.
    \textbf{With a single transformation}, it yields out-of-boundary (OOB) artifacts (such as in {\colorbox{yellow}{yellow boxes}}), as the model has difficulty dealing with both the foreground and the background.
    \textbf{Without $\mathcal{L}_{\mathrm{NSReg}}$}, the model also introduces OOB artifacts.
    \textbf{Without layout augmentation}, the model also predicts distorted results.
    \textbf{With LPIPS loss~}\cite{zhang2018perceptual}, the model predicts distorted results.
    \textbf{Our full model using DreamSim~}\cite{fu2023dreamsim} predicts results with less distortion and avoids OOB artifacts thanks to the compositional transformations.
    }
    \label{fig:qual_ablation}
\end{figure*}
\begin{figure*}[t]    
    \centering
    \captionsetup{type=figure}
    % \mpage{0.02}{\raisebox{50pt}{\rotatebox{90}{Input}}}
    \begin{minipage}[t]{0.195\textwidth}
        \centering
        \frame{\includegraphics[trim=0 0 0 0, clip,height= 0.09\textheight]{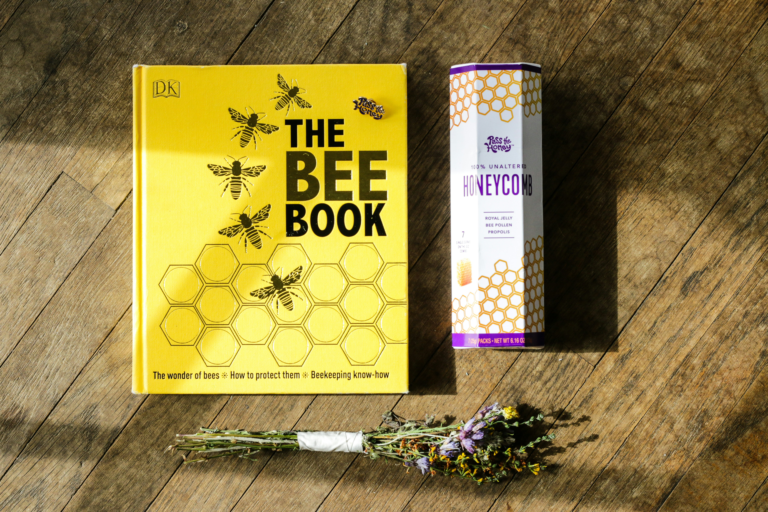}}
    \end{minipage}\hfill 
    \begin{minipage}[t]{0.10\textwidth}
        \centering
        \frame{\includegraphics[trim=0 0 0 0, clip,height= 0.09\textheight]{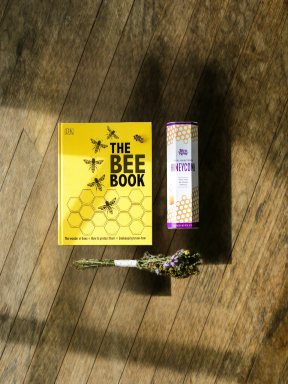}}
    \end{minipage}\hfill 
    \begin{minipage}[t]{0.145\textwidth}
        \centering
        \frame{\includegraphics[trim=0 0 0 0, clip,height= 0.09\textheight]{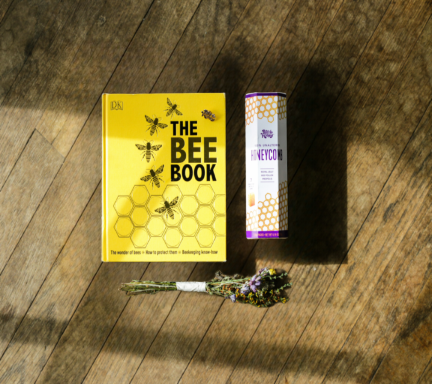}}
        % \caption{}
    \end{minipage}\hfill 
    \begin{minipage}[t]{0.245\textwidth}
        \centering
        \frame{\includegraphics[trim=0 0 0 0, clip,height= 0.09\textheight]{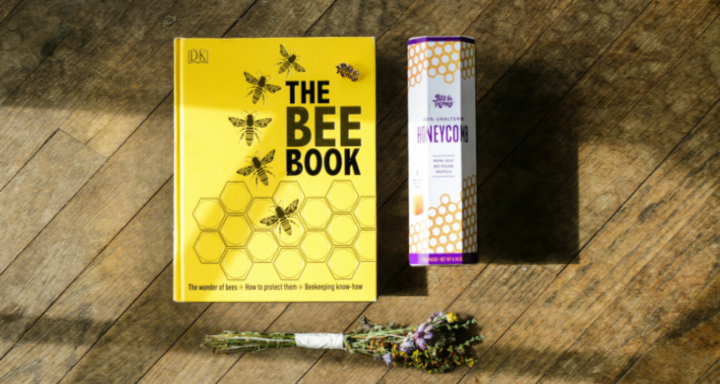}}
        % \caption{}
    \end{minipage}\hfill 
    \begin{minipage}[t]{0.295\textwidth}
        \centering
        \frame{\includegraphics[trim=0 0 0 0, clip,height= 0.09\textheight]{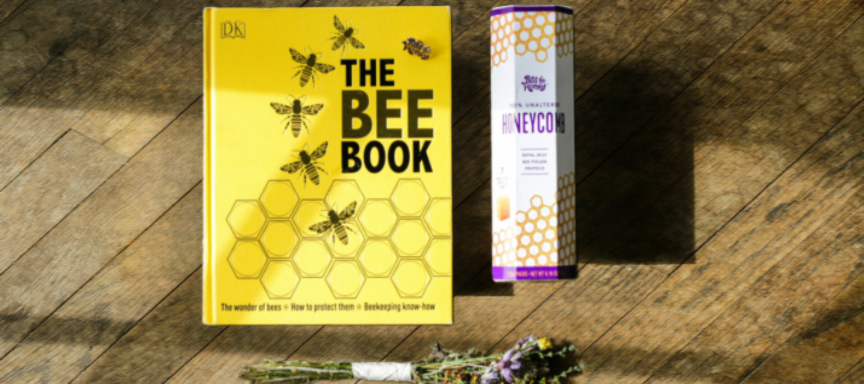}}
        % \caption{}
    \end{minipage}

    % Break to the next row
    % \vspace{1mm} % Adjust the vertical space as needed
    % % \mpage{0.02}{\raisebox{50pt}{\rotatebox{90}{Output}}}
    % \begin{minipage}[t]{0.195\textwidth}
    %     \centering
    %     \frame{\includegraphics[trim=0 0 0 0, clip,height= 0.09\textheight]{images/in_wild_assets/00041.png}}
    % \end{minipage}\hfill 
    % \begin{minipage}[t]{0.10\textwidth}
    %     \centering
    %     \frame{\includegraphics[trim=0 0 0 0, clip,height= 0.09\textheight]{images/in_wild_assets/00041_0.50.png}}
    % \end{minipage}\hfill 
    % \begin{minipage}[t]{0.145\textwidth}
    %     \centering
    %     \frame{\includegraphics[trim=0 0 0 0, clip,height= 0.09\textheight]{images/in_wild_assets/00041_0.75.png}}
    %     % \caption{}
    % \end{minipage}\hfill 
    % \begin{minipage}[t]{0.245\textwidth}
    %     \centering
    %     \frame{\includegraphics[trim=0 0 0 0, clip,height= 0.09\textheight]{images/in_wild_assets/00041_1.25.png}}
    %     % \caption{}
    % \end{minipage}\hfill 
    % \begin{minipage}[t]{0.295\textwidth}
    %     \centering
    %     \frame{\includegraphics[trim=0 0 0 0, clip,height= 0.09\textheight]{images/in_wild_assets/00041_1.50.png}}
    %     % \caption{}
    % \end{minipage}

    \vspace{1mm} % Adjust the vertical space as needed

    \begin{minipage}[t]{0.195\textwidth}
        \centering
        \frame{\includegraphics[trim=0 0 0 0, clip,height= 0.09\textheight]{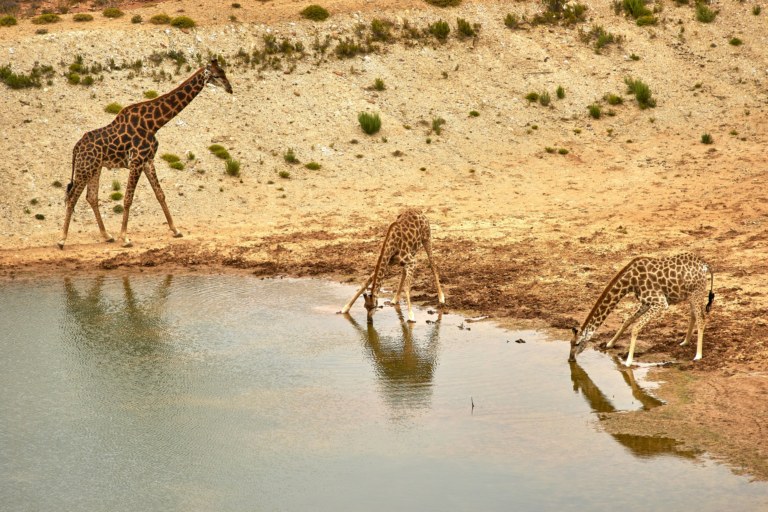}}
        \text{{Input ($H \times W$)}}
    \end{minipage}\hfill 
    \begin{minipage}[t]{0.10\textwidth}
        \centering
        \frame{\includegraphics[trim=0 0 0 0, clip,height= 0.09\textheight]{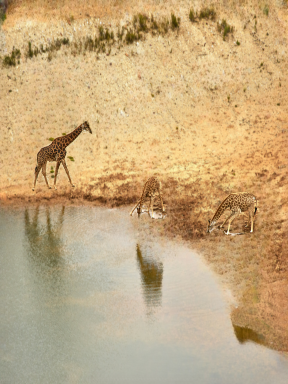}}
        \text{{$0.50W$}}
    \end{minipage}\hfill 
    \begin{minipage}[t]{0.145\textwidth}
        \centering
        \frame{\includegraphics[trim=0 0 0 0, clip,height= 0.09\textheight]{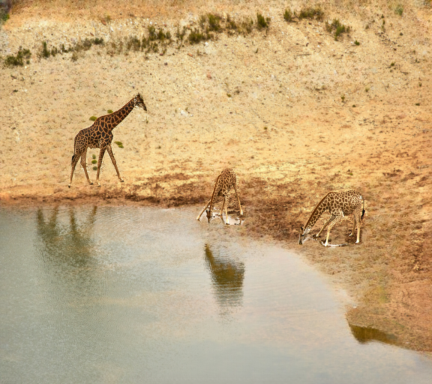}}
        \text{{$0.75W$}}
    \end{minipage}\hfill 
    \begin{minipage}[t]{0.245\textwidth}
        \centering
        \frame{\includegraphics[trim=0 0 0 0, clip,height= 0.09\textheight]{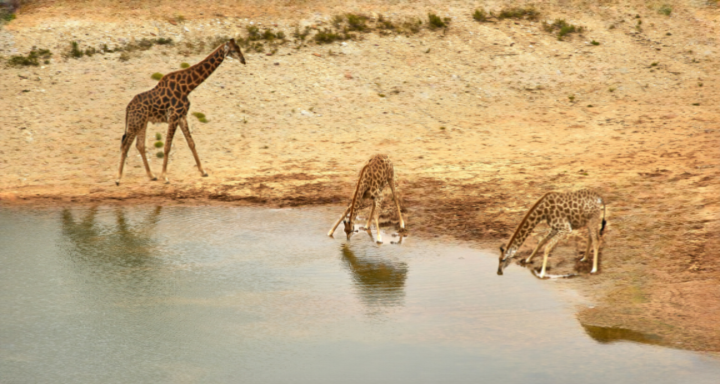}}
        \text{{$1.25W$}}
    \end{minipage}\hfill 
    \begin{minipage}[t]{0.295\textwidth}
        \centering
        \frame{\includegraphics[trim=0 0 0 0, clip,height= 0.09\textheight]{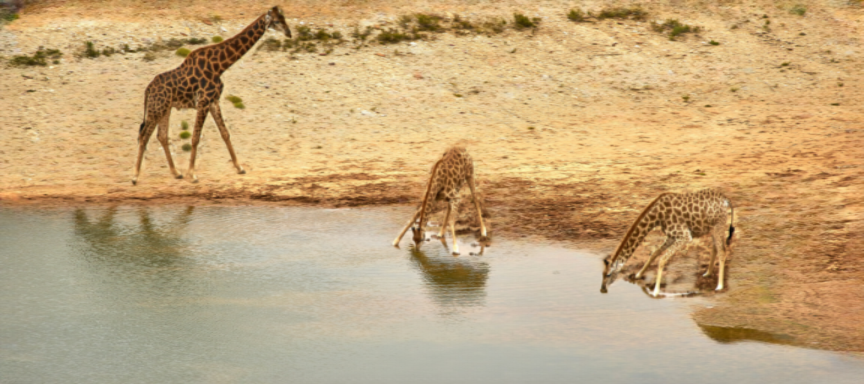}}
        \text{{$1.50W$}}
    \end{minipage}

    \caption{\textbf{Qualitative results on the in-the-wild images.} 
    \emph{Without further finetuning}, our model generalizes to the in-the-wild images, covering common objects and animals. 
    It works for single object and multiple objects.
    The input images are from Unsplash~\cite{unsplash2020unsplash}.
    }
    \label{fig:qual_inwild}
    \vspace{-3mm}
\end{figure*}

\topic{Qualitative comparison.} We showcase some visual comparison in Fig.~\ref{fig:qualitative_comp}. 
We encourage the readers to view the \textbf{Supplementary Material} for more results. 
(a) Compared to optimization-based models~\cite{granot2022drop}, our method better preserves content and structure of the input image, and aesthetically better quality.
(b) Compared to the feed-forward approach, Self-Play-RL~\cite{kajiura2020self}, our method succeeds preserving the content as shown in the ``fish'' and the ``Taj Mahal'' examples.
(c) Compared to DragonDiffsuion~\cite{mou2024dragondiffusion}, our method better preserves the content in ``Taj Mahal'', and shows less visual artifacts in ``fish''.
(d) Our model also emerges with an understanding of ``affordance'', \emph{i.e.}, the ability to place objects appropriately. In the ``fish'' example, our model is the only one successfully positioning the fish behind the sea anemone, maintaining the original spatial relationships.
% \begin{itemize}
%     \item Compared to optimization-based models~\cite{granot2022drop}, our method better preserves content and structure of the input image, and aesthetically better quality.
%     \item Compared to the feed-forward approach, Self-Play-RL~\cite{kajiura2020self}, our method succeeds preserving the content as shown in the ``fish'' and the ``Taj Mahal'' examples.
%     \item Compared to DragonDiffsuion~\cite{mou2024dragondiffusion}, our method better preserves the content in ``Taj Mahal'', and shows less visual artifacts in ``fish''.
%     \item Our model also emerges with an understanding of ``affordance'', \emph{i.e.}, the ability to place objects appropriately. In the ``fish'' example, our model is the only one successfully positioning the fish behind the sea anemone, maintaining the original spatial relationships.
% \end{itemize}
% \yl{I think ours no distortion could benefit from the design of salient mask. We may want to call out the diff here.}
% \yl{Personally I like the term affordance, which indicates a high-level understand of the layout. If we can find more examples, let's put in the supp and mention here.}
% We encourage readers to refer to our \textbf{Supplementary Material} for more comparison results.
% and insights into the model's affordance-awareness.
% \vspace{-6mm}
\begin{figure*}[h!]
    \centering
    \begin{subfigure}[t]{0.28\textwidth}
        \centering
        \frame{\includegraphics[trim=0 0 0 0, clip,height=0.12\textheight]{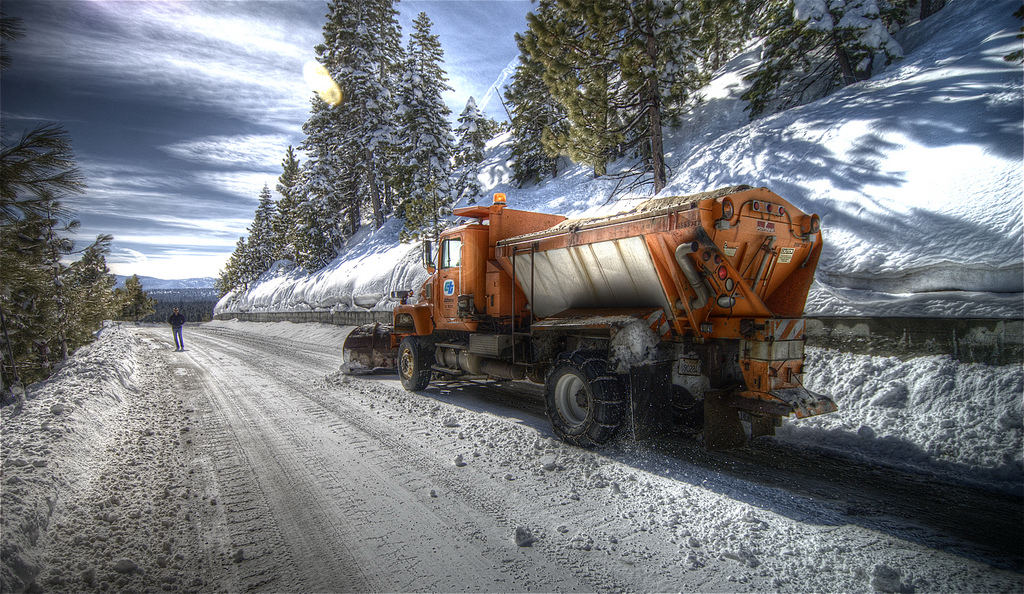}}
        \text{Input }
    \end{subfigure}\hfill
    \begin{subfigure}[t]{0.28\textwidth}
        \centering
        \frame{\includegraphics[trim=0 0 0 0, clip,height=0.12\textheight]{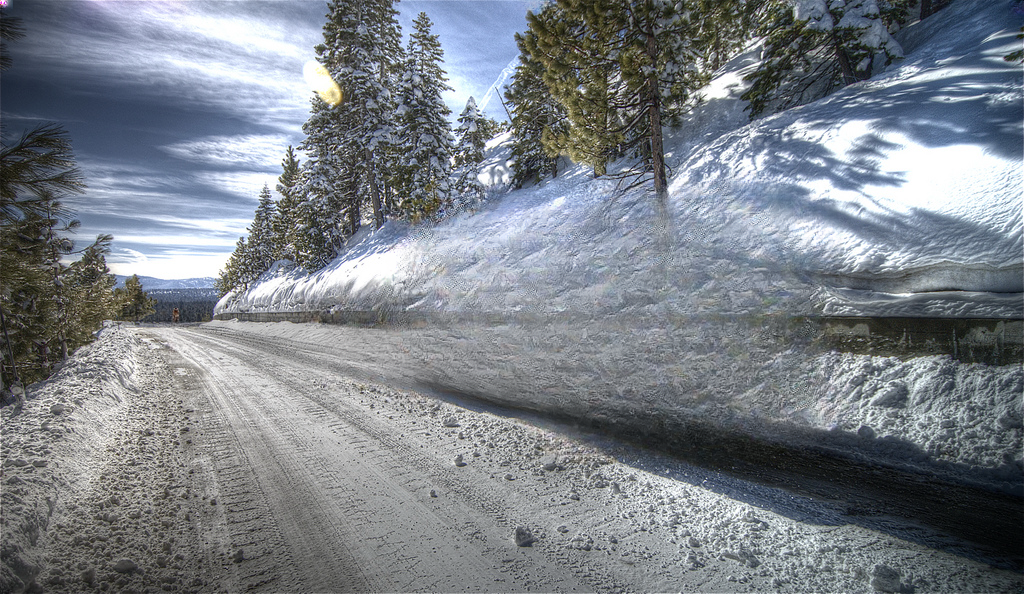}}
        \text{Inpainted by LAMA}
    \end{subfigure}\hfill
    % \begin{subfigure}[t]{0.15\textwidth}
    %     \centering
    %     \frame{\includegraphics[trim=0 0 0 0, clip,height=0.1\textheight]{images/why_lama/LakeVillage_0.75_cv2_ours.png}}
    %     \text{w/ poor SM}
    % \end{subfigure}\hfill
    \begin{subfigure}[t]{0.28\textwidth}
        \centering
        \frame{\includegraphics[trim=0 0 0 0, clip,height=0.12\textheight]{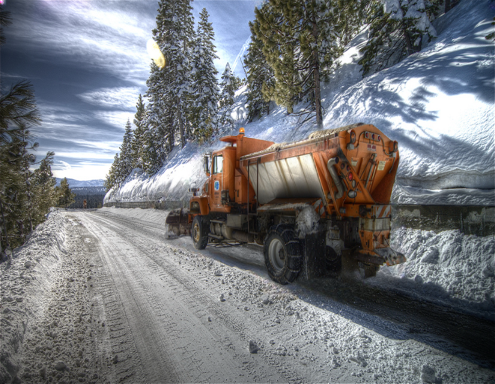}}
        \text{ Our result}
    \end{subfigure}
    % \begin{subfigure}[t]{0.15\textwidth}
    %     \centering
    %     \frame{\includegraphics[trim=0 0 0 0, clip,height=0.1\textheight]{images/why_lama/LakeVillage_ours.png}}
    %     \text{w/ Better SM}
    % \end{subfigure}

    \caption{%\textbf{Affordance helps LAMA.}
    LAMA~\cite{suvorov2022lama} could fail when the inpainting mask is large. 
    In this case, the inpainted result shows undesired textures.
  %   Fortunately, as shown in Figure~\ref{fig:qualitative_comp}, our model emerges with awareness of the affordance. 
 Fortunately, our method places the content correctly and the undesired part is occluded.
    }
    \label{fig:lama_fail}
    \vspace{-3mm}
\end{figure*}
\begin{figure}[h!]
    \centering
    \begin{subfigure}[t]{0.15\textwidth}
        \centering
        \frame{\includegraphics[trim=0 0 0 0, clip,height=0.08\textheight]{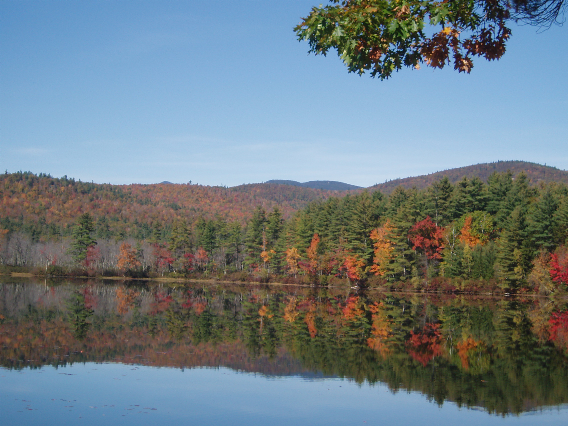}}
        \text{\small{Input} }
    \end{subfigure}\hfill
    \begin{subfigure}[t]{0.15\textwidth}
        \centering
        \frame{\includegraphics[trim=0 0 0 0, clip,height=0.08\textheight]{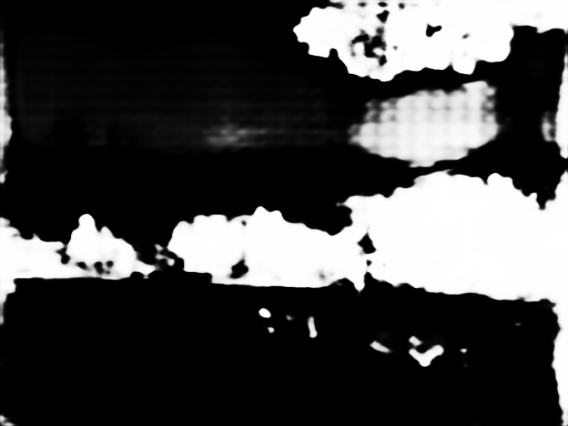}}
        \text{\small{MDSAM prediction}}
    \end{subfigure}\hfill
    % \begin{subfigure}[t]{0.22\textwidth}
    %     \centering
    %     \frame{\includegraphics[trim=0 0 0 0, clip,height=0.11\textheight]{images/mdsam_fail/foliage_0.75_ours.png}}
    %     \text{Result w/ MDSAM}
    % \end{subfigure}\hfill
    \begin{subfigure}[t]{0.15\textwidth}
        \centering
        \frame{\includegraphics[trim=0 0 0 0, clip,height=0.08\textheight]{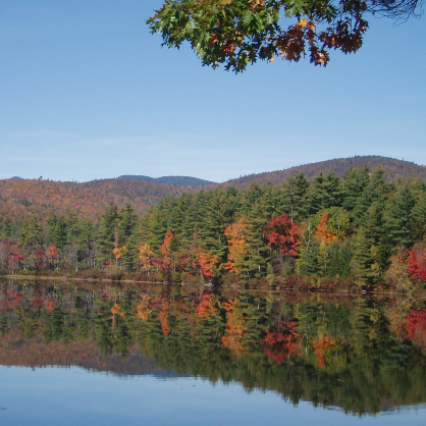}}
        \text{\small{Ours w/ all-one mask}}
    \end{subfigure}

    \caption{\textbf{All-one mask helps MDSAM.}
  When there are no obvious salient objects, it may produce unreliable results. In that case, we can provide the model with an \textbf{all-one} mask, and our model becomes a cropping model.
    }
    \label{fig:mdsam_fail}
    \vspace{-3mm}
\end{figure}
\begin{table}[h!]
  \centering
    \caption{\textbf{Ablation study.}
      We study the effect of different components. All scores are shown in percentage  (\%). With a single transformation, the model achieves a lower DreamSim error, yet it has OOB issue as shown in Figure~\ref{fig:qual_ablation}. 
      }
    \resizebox{0.48\textwidth}{!}{
    \begin{tabular}{lcccc}
    \toprule
    %       & \multicolumn{1}{c}{Content} & \multicolumn{1}{c}{Layout} & \multicolumn{1}{c}{Aesthetics}  \\
    % \midrule
          & \multicolumn{1}{c}{CLIP$\uparrow$} & \multicolumn{1}{c}{DreamSim$\uparrow$} & \multicolumn{1}{c}{MUSIQ$\uparrow$}  & \multicolumn{1}{c}{Average}\\
    \midrule
    Single Transformation   &  88.3  &  \textbf{80.8}  & 47.9 & 72.3 \\
    w/o $\mathcal{L}_{\mathrm{NSReg}}$          &  83.6  & 77.3   & 45.3 & 68.7   \\
    % w/o augmentation     &  \underline{89.69}  & 76.9   & 48.9  \\
    % % \midrule               
    % Ours (w/ LPIPS)         &  89.67  &  76.9  & \underline{49.2}      \\
    % Ours (w/ DreamSim)            &   \textbf{90.17}  &  \underline{78.1} &  \textbf{51.5} \\
    Ours (LPIPS)         &  89.7  &  76.9  & {49.2}  & 71.9    \\
    Ours (LPIPS + aug.)         &  \textbf{90.2}  &  {77.2}\underline  & \underline{50.3}    & \underline{72.6}  \\
    Ours (DreamSim)     &  {89.7}  & 76.9   & 48.9  & 71.8  \\
    Ours (DreamSim + aug.)            &   \textbf{90.2}  &  \underline{78.1} &  \textbf{51.5} & \textbf{73.3} \\
    \bottomrule
    \end{tabular}%
    }
  \vspace{-5mm}
  \label{tab:ablation}%
\end{table}%
\subsection{Ablation study}
To examine the effect of each proposed component, we conduct an ablation study. 
We remove one component in our full method each time, and show the results in Tab.~\ref{tab:ablation} and Fig.~\ref{fig:qual_ablation}. \textbf{With one single transformation}, the model achieves the best performance on structure preservation, but it introduces OOB artifacts as shown in Fig.~\ref{fig:qual_ablation}. \textbf{Removing the background regularization term} $\mathcal{L}_{\mathrm{NSReg}}$ also introduces some OOB artifacts as shown in Fig.~\ref{fig:qual_ablation}. \textbf{Layout augmentation} brings significant improvement for the distortions, as shown in Tab.~\ref{tab:ablation} and Fig.~\ref{fig:qual_ablation}. Finally, by \textbf{replacing DreamSim with LPIPS}~\cite{zhang2018perceptual}, the model still suffers from the distorted content, further highlighting DreamSim's effectiveness in maintaining layout and structure awareness.

\subsection{Analysis of the off-the-shelf models}
We use two off-the-shelf models, LAMA~\cite{suvorov2022lama} for the inpainting and MDSAM~\cite{gao2024multiscale} for the saliency detection. We now provide solutions if these methods fail.

% \topic{Inpainting model.}LAMA~\cite{suvorov2022lama} could compromise when the textures are complicated. Fortunately, as shown in Fig.~\ref{fig:qualitative_comp},  our model emerges with awareness of the affordance.  It, therefore, places the content correctly, and the undesired part is occluded. We show an example in Fig.~\ref{fig:lama_fail}.
\topic{Inpainting.} LAMA struggles with complex textures, but our model, %aware of affordance, 
correctly places content while occluding undesired regions (Fig.~\ref{fig:lama_fail}).

% \topic{Saliency detector.}When there are no obvious salient objects, MDSAM may produce unreliable results. In that case, we can provide the model with an all-one mask, and our model becomes a cropping model. We show an example in Fig.~\ref{fig:mdsam_fail}. 
% We would like to emphasize that, for this challenging case (no obvious saliency), it is ill-posed and there are multiple solutions. 
\topic{Saliency Detection.} When MDSAM fails due to a lack of salient objects, we use an all-one mask, turning our model into a cropping-based approach (Fig.~\ref{fig:mdsam_fail}). Notably, such cases are inherently ill-posed with multiple valid solutions.

\subsection{Results on In-the-wild data}
To demonstrate the generalizability of our model, we test our model on 400 in-the-wild images from Unsplash~\cite{unsplash2020unsplash}.
% We show qualitative results in Fig.~\ref{fig:qual_inwild}.
% \emph{Without any finetuning}, our model generalizes well to diverse scenarios, varying from common objects, natural landscapes, and animals.
We show results on in-the-wild data \emph{without any finetuning} in Fig.~\ref{fig:qual_inwild} and in the \textbf{Supplementary Material}.

\subsection{Discussions}
\topic{Limitations.} \methodname also has limitations from content associations (\eg, ball and its shadow). We show an example and a solution in the supplementary material.

\topic{Social impacts.} 
\methodname preserves the content of the input image, minimizing potential negative social impacts.

% \subsection{Limitations}
% \yiran{Feel free to move this part to the supp.}
% Our current approach also has limitations. 
% As shown in Fig.~\ref{fig:limitations}, \methodname struggles when the saliency detector~\cite{gao2024multiscale} fails to associate the soccer ball with its shadow.
% We can either use a more accurate mask (\textit{e.g.,} from~\cite{liu2023grounding}) or use an object association method~\cite{alzayer2024magic,winter2024objectdrop} to improve the result.
% \input{figures/fig7_limitations}
% \yl{to be more accurate, is this limitation more like our dependency limitation or our design of the method? The explanation reads a bit mixed to me.}
% \yiran{I feel it's more like the dependency limitation as the results can be improved with simply replacing a better saliency mask}
\section{Conclusion}\label{sec:conclusion}
We present \methodname, an end-to-end framework for image retargeting that aligns with human perception.
By using a layered representation for the input and applying distinct transformations to salient and non-salient regions, our approach produces results with fewer visual artifacts, such as the OOB issue.
We also introduce a new Perceptual Structure Similarity Loss (PSSL) enabling training without paired data for image retargeting and equips the model with distortion-awareness capabilities.
We conduct extensive evaluations across various methods, demonstrating that \methodname outperforms previous approaches. 
A user study further confirms that \methodname aligns closely with human perception, outperforming the SOTAs by a large margin.

%We finally discuss the limitations of \methodname and propose several directions for the future work.

% {
%     \small
%     \bibliographystyle{ieeenat_fullname}
%     \bibliography{main}
% }
{\small
\bibliographystyle{ieee_fullname}
\bibliography{main}
}

\newpage
\onecolumn
\appendix
\section{Supplementary Material}\label{sec:appendix}
\subsection{Implementation details}
\subsubsection{Network architecture}
\textbf{Encoder.} We use the same encoder as in GANGealing~\cite{peebles2022gg}.
The encoder follows the architecture of the StyleGAN2 discriminator~\cite{Karras2019stylegan2}, with a ResNet backbone~\cite{he2016residual}. 
In practice, we use the same encoder for both the original input image and its naively resized version to obtain two feature maps.
Two feature maps are fed into $L$ Cross-Attention blocks.

\textbf{Cross-Attention blocks.} To condition the network on the target image size, we choose to naively resize the input image to the target size. 
To better understand the rough layout at the target size, and to introduce a differentiable analogy to PNN methods~\cite{granot2022drop,elnekave2022generating}, we choose to use cross-attention mechanism to share the information between the original input and the resized input.
We adopt the decoder block from CroCo-v2~\cite{croco_v2}, where it consists of $\mathrm{LayerNorm}$, $\mathrm{SelfAttention}$, $\mathrm{CrossAttention}$ and $\mathrm{MLP}$.
In practice, we use $L=3$ decoder blocks, and each block has $4$ heads.

\textbf{Heads.} We use two heads for the foreground and the background, respectively.
Each head predicts an affine transformation.
Unlike GANGealing~\cite{peebles2022gg} and NeuralGealing~\cite{ofri2023neural}, which compose a similarity transformation with an unconstrained flow field, we find the flow field introduces unnatural distortions so we end up without using the flow field.
In practice, each head is equipped with a $\mathrm{Linear}$ layer to predict 5 parameters $o_1,o_2,o_3,o_4,o_5$. 
We construct the affine matrix $\rmA$ as follows:
\begin{align}
    r &= \pi \cdot \tanh{(o_1)}  \\
    s_x &= \exp{(o_2)}  \\
    s_y &= \exp{(o_3)} \\
    t_x &= o_4  \\
    t_y &= o_5 \\
    \rmA &= \begin{bmatrix}
            s_x \cdot \cos{(r)} & -s_y \cdot \sin{(r)} & t_x \\
            s_y \cdot \sin{(r)} &  s_x \cdot \cos{(r)} & t_y  \\
            0 & 0 & 1 
            \end{bmatrix}
\end{align}
% \begin{equation}
%     s_x = \exp{(o_2)}
% \end{equation}\vspace{-4mm}
% \begin{equation}
%     s_y = \exp{(o_3)}
% \end{equation}\vspace{-4mm}
% \begin{equation}
%     t_x = o_4
% \end{equation}\vspace{-4mm}
% \begin{equation}
%     t_y = o_5
% \end{equation}\vspace{-4mm}
% \begin{equation}
%     \rmA = \begin{bmatrix}
%             s_x \cdot \cos{(r)} & -s_y \cdot \sin{(r)} & t_x \\
%             s_y \cdot \sin{(r)} &  s_x \cdot \cos{(r)} & t_y  \\
%             0 & 0 & 1 
%             \end{bmatrix}
% \end{equation}
To warp the image, we apply $\rmA$ to an identity sampling grid, and then apply the transformed sampling grid to the input image.

% \subsubsection{Layout disturbance}
\subsubsection{Perceptual structure similarity loss}
We apply a random transformation to the input image as an undistorted, pseudo ground truth during the training.
The transformation includes a scaling ${s}$ and a translation $\mathbf{t} = [t_1, t_2] \in \mathbb{R}^2$.
Suppose the input image has a size of $H \times W$, and the target size is $H^{\prime} \times W^{\prime}$,
we construct a transformation matrix $\rmD$ as follows:
\begin{equation}
    \rmD = \begin{bmatrix}
            s \cdot k_x  & -s \cdot k_y  & t_1 \cdot H^{\prime} \\
            s \cdot k_y  &  s \cdot k_x  & t_2 \cdot W^{\prime} \\
            0 & 0 & 1 
    \end{bmatrix} \,,
\end{equation}
where $k_x = \frac{H^{\prime}}{H}$, $k_y = \frac{W^{\prime}}{W}$. 
To obtain the warped image, we apply $\rmD$ to an identity sampling grid, and then apply the transformed sampling grid to the input image. 
In practice, we sample $s$ from a uniform distribution $\mathcal{U} \sim [0.9, 1.5]$ and $t_1, t_2$ from  $\mathcal{U} \sim [-0.01, 0.01]$.

\begin{figure}[h!]
    \centering
    \begin{subfigure}[t]{0.25\textwidth}
        \centering
        \frame{\includegraphics[trim=0 0 0 0, clip,height=0.12\textheight]{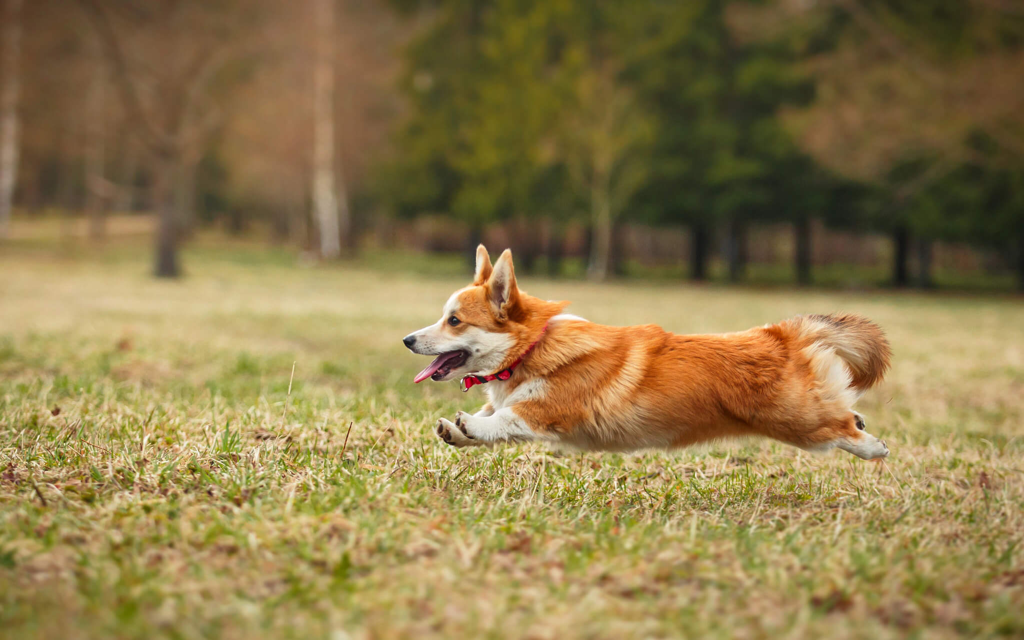}}
        \text{Input}
    \end{subfigure} \hspace{15mm}
    \begin{subfigure}[t]{0.20\textwidth}
        \centering
        \frame{\includegraphics[trim=0 0 0 0, clip,height=0.12\textheight]{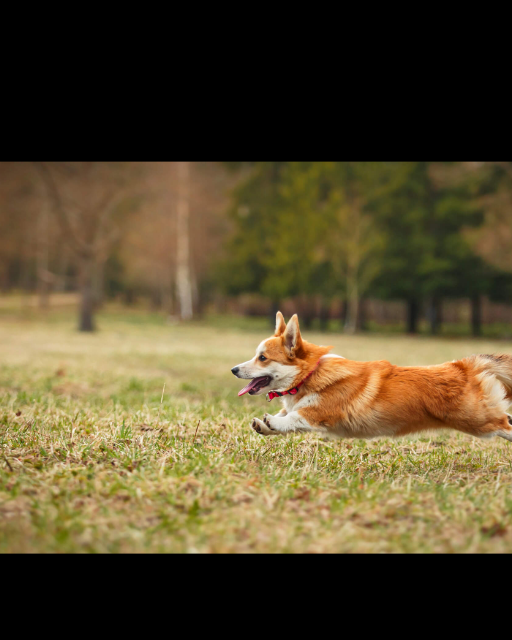}}
        \text{Example 1}
    \end{subfigure} \hfill
    \begin{subfigure}[t]{0.20\textwidth}
        \centering
        \frame{\includegraphics[trim=0 0 0 0, clip,height=0.12\textheight]{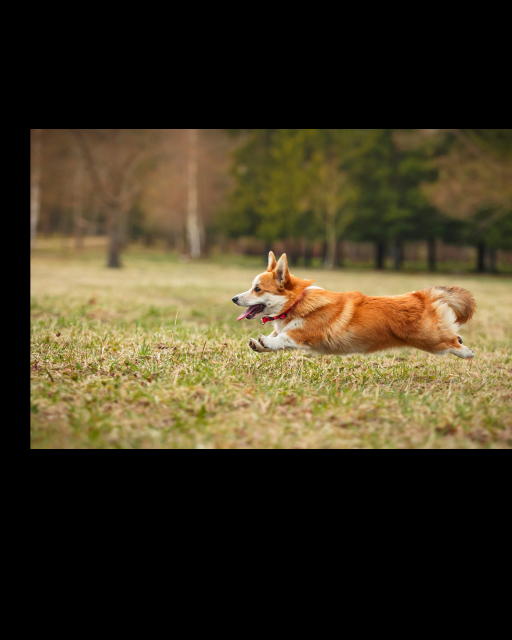}}
        \text{Example 2}
    \end{subfigure} \hfill
    \begin{subfigure}[t]{0.20\textwidth}
        \centering
        \frame{\includegraphics[trim=0 0 0 0, clip,height=0.12\textheight]{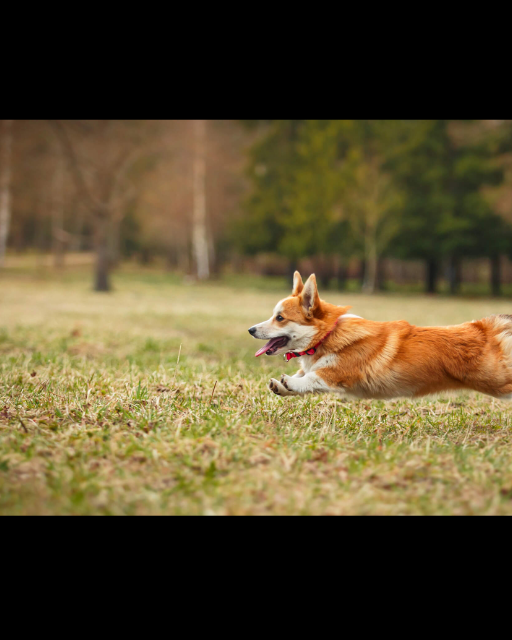}}
        \text{Example 3}
    \end{subfigure}
    % \begin{subfigure}[t]{0.20\textwidth}
    %     \centering
    %     \frame{\includegraphics[trim=0 0 0 0, clip,height=0.12\textheight]{images/layout_disturb_assests/image_random_trans_iter_4000.png}}
    % \end{subfigure}

    % \mpage{0.15}{{Input}}\hfill
    % \mpage{0.15}{{GPNN~\cite{granot2022drop}}}\hfill
    % \mpage{0.15}{{Self-Play-RL~\cite{kajiura2020self}}}
    
    \caption{\textbf{Examples of layout augmentation.}
    We show some examples of the layout augmentation. In this case, $H^{\prime}=0.5H$.
    }
    \label{fig:layout_disturb}
\end{figure}

We show some examples of the random augmented images in Fig.~\ref{fig:layout_disturb}.

\subsubsection{Training details}
We train our network with an initial learning rate $\alpha=1 \times 10^{-4}$ and an Adam optimizer~\cite{adam}. The learning rate decays by a factor of $0.9$ in every 1000 iterations.
To facilitate batch training, we split the images with same aspect-ratio into different groups.
At each iteration, we sample a group and a batch of images from the group.
We use a batch size of 32 and train the model for 200 epochs.
During the training, we sample a random target ratio from $\{0.50, 0.75, 1.25, 1.50\}$ at each iteration. We then randomly choose to scale the height or the width of the image with the sampled ratio factor for current batch.
We train our model with 2 NVIDIA A100 GPUs for around 2 days.

\subsection{User study}
We conduct the user study using Mturk~\cite{crowston2012amazon}. On the user study interface, we place the original image at the top and the output of our method and the other method on the same line side by side, one at a time for each image. 
We randomly assign the left/right position for each image to eliminate positional bias from users. We compare our method with the following methods: Self-Play-RL~\cite{kajiura2020self},  GPNN\cite{granot2022drop}, and DragonDiffusion\cite{mou2024dragondiffusion}. 
We do aspect ratio preserving resizing for all images from different methods to match the same resolution for HTML Compatibility. Uniform image dimensions enhance compatibility and visual coherence when displaying results in HTML files, leading to a more seamless and professional presentation.
Each image is voted by 5 independent evaluators.

\begin{figure}[htbp]
  \centering
  \includegraphics[width=\textwidth]{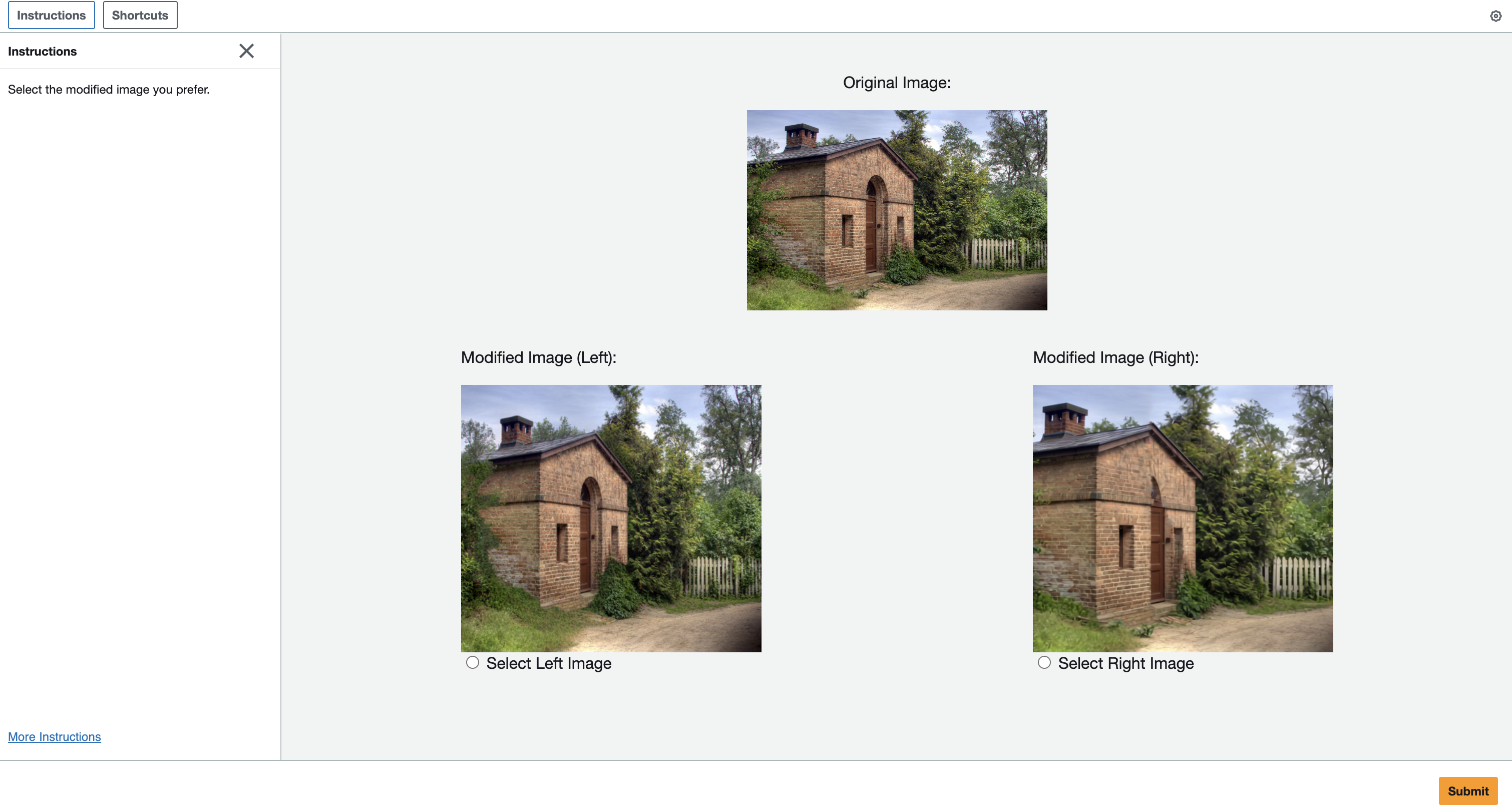}
  \caption{User Study Interface}
  \label{fig:screenshot1_vertical}
\end{figure}

Fig.~\ref{fig:screenshot1_vertical}  is an example of the user study interface for users. 

%\begin{figure}[htbp]
%  \centering
%  \includegraphics[width=0.7\textwidth]{images/instruct2.png}
%  \caption{Instruction for Users}
%  \label{fig:screenshot2_vertical}

%  \caption{User Study Example for our method VS other methods}
%  \label{fig:two_screenshots_vertical}
%\end{figure}

We give the following instruction to users:

\begin{enumerate}
    \item View the original image at the top.
    \item Two modified versions are shown below. Select the one you like better.
    \item Choose the `Select Left Image' or `Select Right Image' option below the corresponding modified image.
    \item We evaluate an image by content, structure, and artifact.
    \begin{itemize}
        \item \textbf{Content:}
            \begin{itemize}
                \item The result with CROPPED or MISSING part is worse.
                \item The modified image should still show the main content of the original image without adding or reducing some of the content.
            \end{itemize}
        \item \textbf{Structure:}
            \begin{itemize}
                \item The modified image shows correct structure for each object, with less distortions.
            \end{itemize}
        \item \textbf{Artifact:}
            \begin{itemize}
                \item The modified image should have less visual artifacts, e.g., cut-off objects, weird structures.
            \end{itemize}
    \end{itemize}
\end{enumerate}

All 80 images in RetargetMe~\cite{rubinstein2010retargetme} are evaluated. After getting the vote of 5 users on each image, we select the method that at least 3 users choose to be the winner for each image and calculate the win-rate of our approach VS the others. 

\begin{figure*}[t]
    \centering
    % \scriptsize
    \begin{subfigure}[t]{0.25\textwidth}
        \centering
        \frame{\includegraphics[trim=0 0 0 0, clip,height=0.103\textheight]{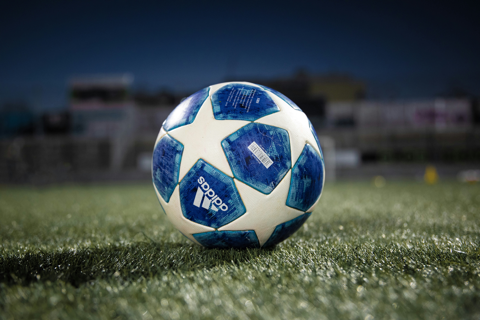}}
        \text{ Input}
    \end{subfigure}\hfill
    \begin{subfigure}[t]{0.25\textwidth}
        \centering
        \frame{\includegraphics[trim=0 0 0 0, clip,height=0.103\textheight]{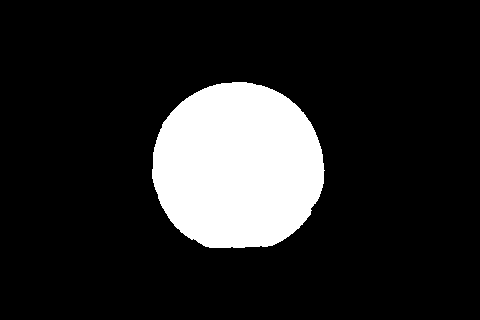}}
        \text{Poor SM}
    \end{subfigure}\hfill
    \begin{subfigure}[t]{0.125\textwidth}
        \centering
        \frame{\includegraphics[trim=0 0 0 0, clip,height=0.103\textheight]{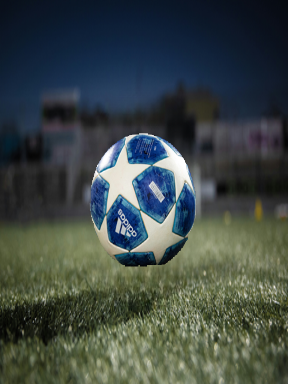}}
        \text{w/ poor SM}
    \end{subfigure}\hfill
    \begin{subfigure}[t]{0.25\textwidth}
        \centering
        \frame{\includegraphics[trim=0 0 0 0, clip,height=0.103\textheight]{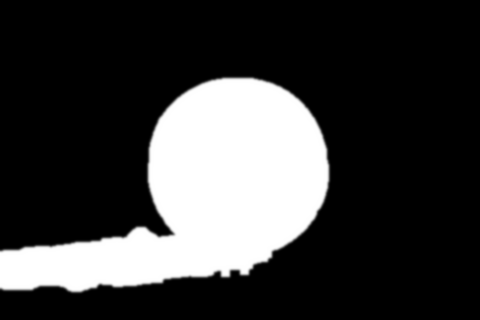}}
        \text{Better SM}
    \end{subfigure}\hfill
    \begin{subfigure}[t]{0.125\textwidth}
        \centering
        \frame{\includegraphics[trim=0 0 0 0, clip,height=0.103\textheight]{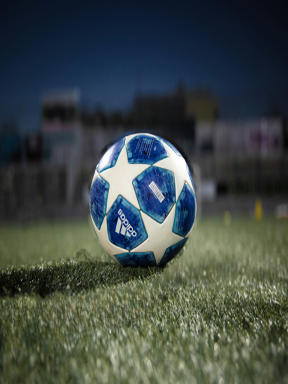}}
        \text{w/ Better SM}
    \end{subfigure}

    \caption{\textbf{Limitations.}
    Our model faces challenges with poor saliency map (SM) prediction. In this example, the saliency detector of~\cite{gao2024multiscale} fails to associate the shadow with the soccer ball, resulting in a ``floating'' ball.
    By using an improved mask that includes the shadow, our model yields a more reasonable output.
    We reduce the width of the image to its half in this case.
    }
    \label{fig:limitations}
\end{figure*}
\subsection{Limitations}
% \yiran{Feel free to move this part to the supp.}
Our current approach also has limitations. 
As shown in Fig.~\ref{fig:limitations}, \methodname struggles when the saliency detector~\cite{gao2024multiscale} fails to associate the soccer ball with its shadow.
We can either use a more accurate mask (\textit{e.g.,} from~\cite{liu2023grounding}) or use an object association method~\cite{alzayer2024magic,winter2024objectdrop} to improve the result.

\subsection{Analysis of the off-the-shelf models}
\subsubsection{Analysis of the inpainting model}
We use an off-the-shelf inpainting model, LAMA~\cite{suvorov2022lama}, one of the state-of-the-art image inpainting models.

\textbf{Why LAMA?} We show a qualitative comparison with another naive inpainting method from OpenCV library in Fig.~\ref{fig:why_lama}. 

\textbf{If LAMA fails.} As an off-the-shelf model, LAMA could compromise when the textures are complicated. Fortunately, as shown in Fig.~\ref{fig:supp_affordance}, our model emerges with awareness of the affordance. It therefore places the content correctly and the undesired part is occluded. 
% We show an example in Fig.~\ref{fig:lama_fail}.
% \input{figures/fig_rebuttal_lama_fail}

\subsubsection{Analysis of the saliency detector}
We use one of the state-of-the-art saliency detectors, MDSAM~\cite{gao2024multiscale} to predict saliency map.

\textbf{Why MDSAM?} To demonstrate the effectiveness of MDSAM, we retrain a model without MDSAM and use an all-one mask instead.
We show the performance in Table~\ref{tab:no_saliency}.
Without saliency detector~\cite{gao2024multiscale}, the model shows a similar result as Single Transformation. It shows a higher DreamSim as the preprocessing of DreamSim prefers a distorted result (Fig.~\ref{fig:layout_disturb}). Our full model shows the highest average score over three metrics.
\begin{table}[t]
  \centering
%   \scriptsize
    \caption{\textbf{Performance without saliency detector.}
     Without saliency detector~\cite{gao2024multiscale}, the model shows a similar result as Single Transformation. It shows a higher DreamSim as the preprocessing of DreamSim prefers a distorted result (Figure~\ref{fig:layout_disturb}). Our full model shows the highest average score over three metrics.
      }
    \begin{tabular}{lcccc}
    \toprule
    %       & \multicolumn{1}{c}{Content} & \multicolumn{1}{c}{Layout} & \multicolumn{1}{c}{Aesthetics}  \\
    % \midrule
          & \multicolumn{1}{c}{CLIP sim.(\%)$\uparrow$} & \multicolumn{1}{c}{DreamSim sim.(\%)$\uparrow$} & \multicolumn{1}{c}{MUSIQ(\%)$\uparrow$} & average  \\
    \midrule
    Single Transformation   &  88.33  &  \textbf{80.8}  & 47.9 & 72.3  \\
    w/o saliency detector          &  87.25  & 82.1   & 48.6 & 72.6  \\
    Ours (full)            &   \textbf{90.17}  &  \underline{78.1} &  \textbf{51.5} & \textbf{73.3} \\
    \bottomrule
    \end{tabular}%
  
%   \vspace{-5mm}
  \label{tab:no_saliency}%
\end{table}%

\textbf{If MDSAM fails.} When there are no obvious salient objects, MDSAM may produce unreliable results. In that case, we can provide the model with an all-one mask, and our model becomes a cropping model. 
% We show an example in Fig.~\ref{fig:mdsam_fail}. 
We would like to emphasize that, for this challenging case (no obvious saliency), it is ill-posed and there are multiple solutions. 

\subsection{Limitation of MUSIQ score}
We find MUSIQ~\cite{ke2021musiq} sometimes prefers results with distortions. We show an example in Fig.~\ref{fig:musiq_limit}.

\subsection{Additional results}
\subsubsection{Additional results for affordance-awareness}
As mentioned in Figure 6 in our main paper, our model emerges an ability to understand the affordance between different objects in Fig.~\ref{fig:supp_affordance}.
We show more results to demonstrate the understanding of affordance.

\subsubsection{Additional comparison with previous methods}
\begin{figure*}[h!]    
    \centering
    \includegraphics[trim=0 0 0 0, clip,width=\textwidth]{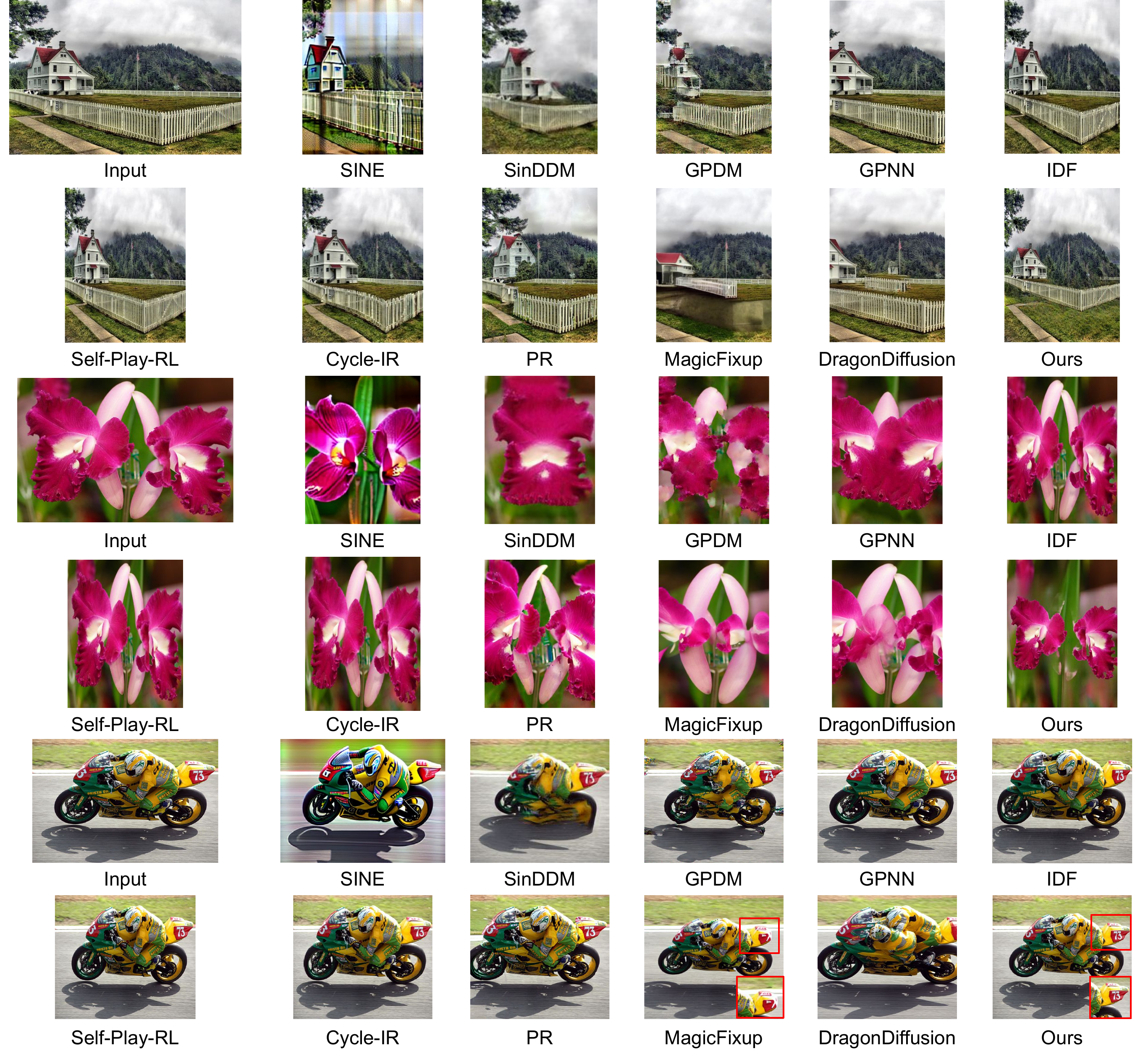}
    \caption{\textbf{Additional qualitative comparison on RetargetMe.} 
    We show more visual comparison results. 
    We compare with SINE~\cite{zhang2023sine}, SinDDM~\cite{kulikov2023sinddm}, GPDM~\cite{elnekave2022generating}, GPNN~\cite{granot2022drop}, IDF~\cite{elsner2024retargeting}, PR~\cite{shen2024prune}, Self-Play-RL~\cite{kajiura2020self}, Cycle-IR~\cite{tan2019cycle}, WSSDCNN~\cite{cho2017wssdcnn}, MagicFixup~\cite{alzayer2024magic}, DragonDiffusion~\cite{mou2024dragondiffusion}
    }
    \label{fig:supp_qual_comp_1}
\end{figure*}

\begin{figure*}[h!]     
    \centering
    \includegraphics[trim=0 0 0 0, clip,width=\textwidth]{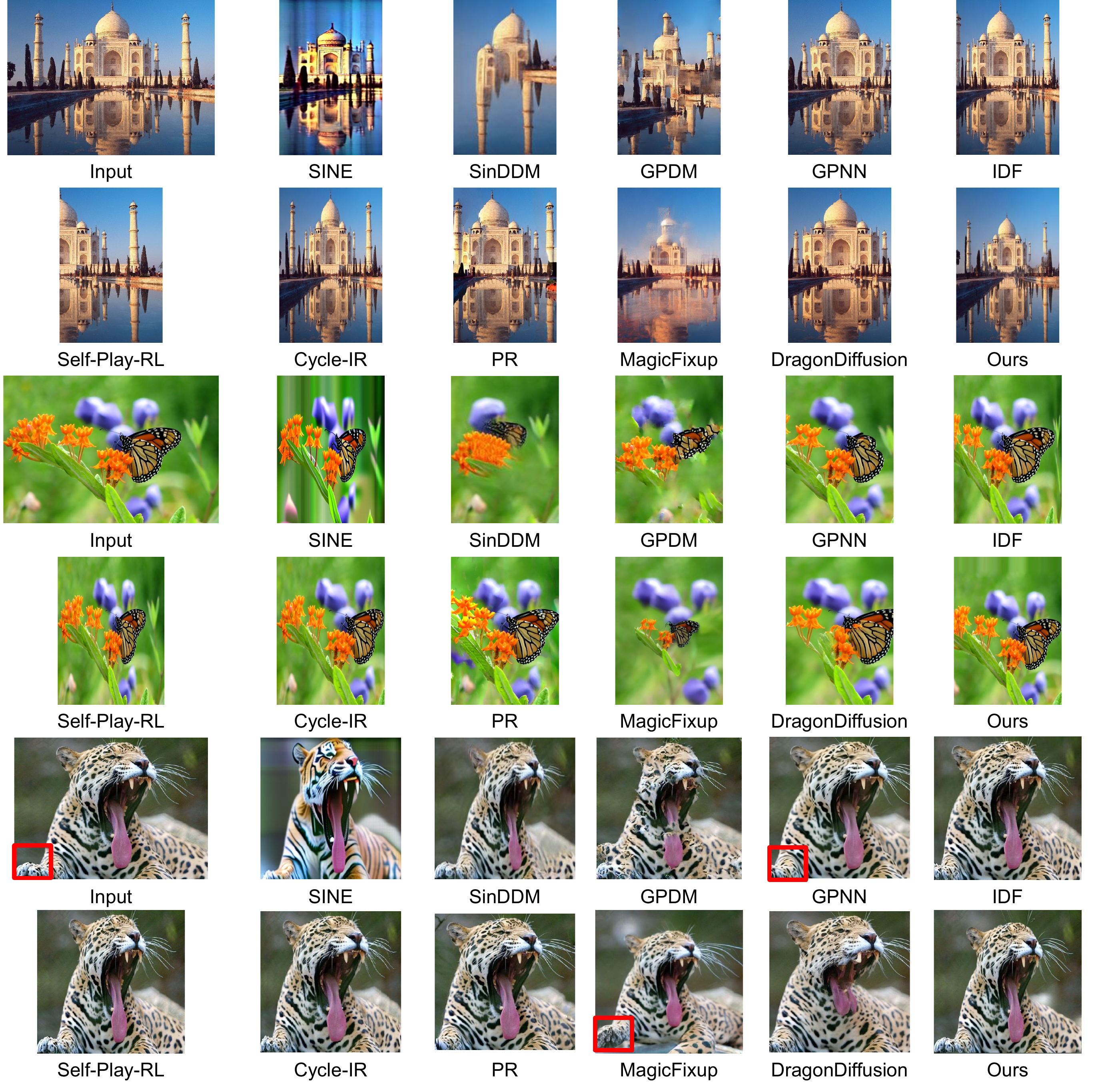}
    \caption{\textbf{Additional qualitative comparison on RetargetMe.} 
    We show more visual comparison results. 
    We compare with SINE~\cite{zhang2023sine}, SinDDM~\cite{kulikov2023sinddm}, GPDM~\cite{elnekave2022generating}, GPNN~\cite{granot2022drop}, IDF~\cite{elsner2024retargeting}, PR~\cite{shen2024prune}, Self-Play-RL~\cite{kajiura2020self}, Cycle-IR~\cite{tan2019cycle}, WSSDCNN~\cite{cho2017wssdcnn}, MagicFixup~\cite{alzayer2024magic}, DragonDiffusion~\cite{mou2024dragondiffusion}.
    }
    \label{fig:supp_qual_comp_3}
\end{figure*}

\begin{figure}[h!]
    \centering
    \begin{subfigure}[t]{0.30\textwidth}
        \centering
        \frame{\includegraphics[trim=0 0 0 0, clip,width=\textwidth]{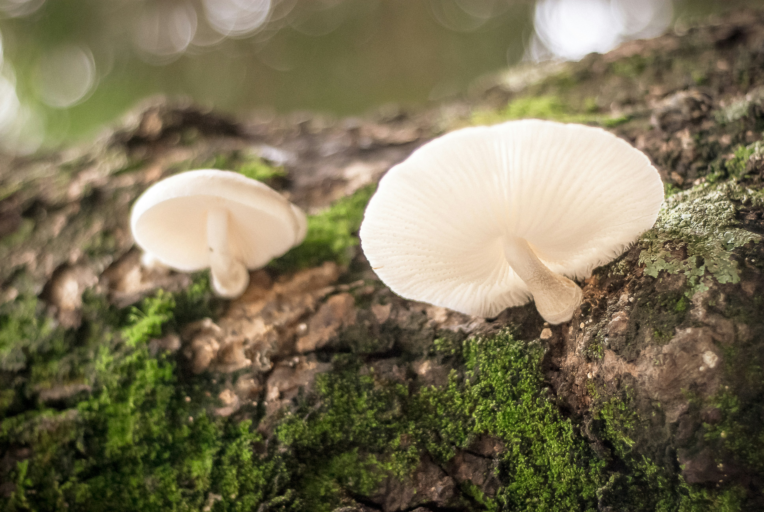}}
    \end{subfigure} 
    \hspace{3mm}
    \begin{subfigure}[t]{0.45\textwidth}
        \centering
        \frame{\includegraphics[trim=0 0 0 0, clip,width=\textwidth]{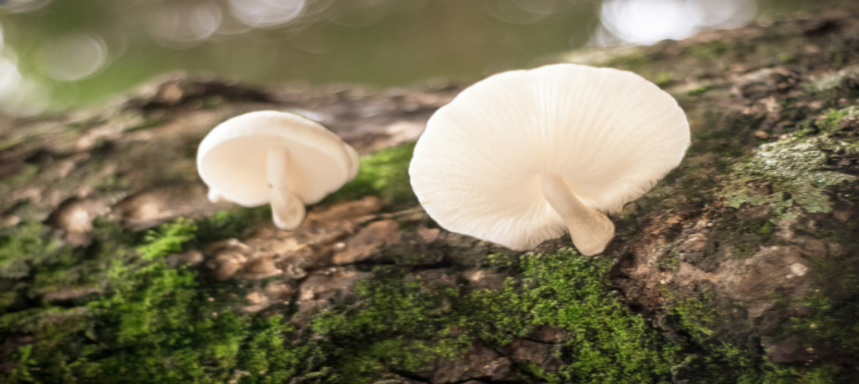}}
    \end{subfigure}
    \vspace{3mm}
    
    \begin{subfigure}[t]{0.30\textwidth}
        \centering
        \frame{\includegraphics[trim=0 0 0 0, clip,width=\textwidth]{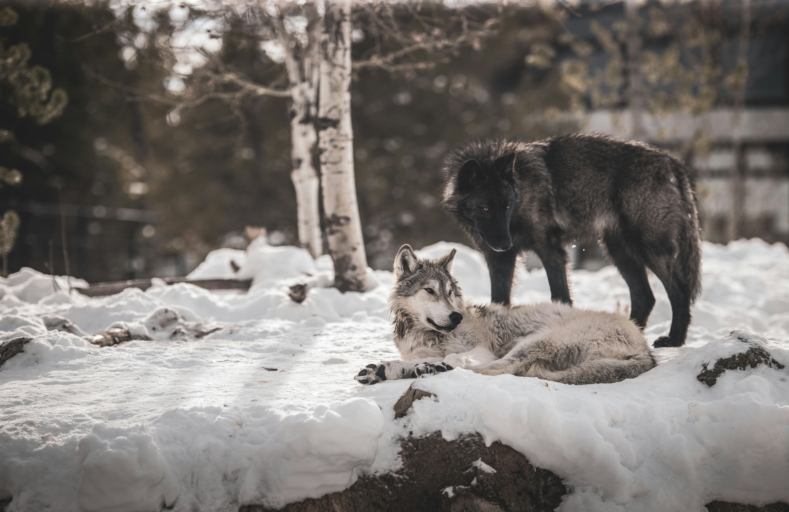}}
    \end{subfigure}
    \hspace{3mm}
    \begin{subfigure}[t]{0.45\textwidth}
        \centering
        \frame{\includegraphics[trim=0 0 0 0, clip,width=\textwidth]{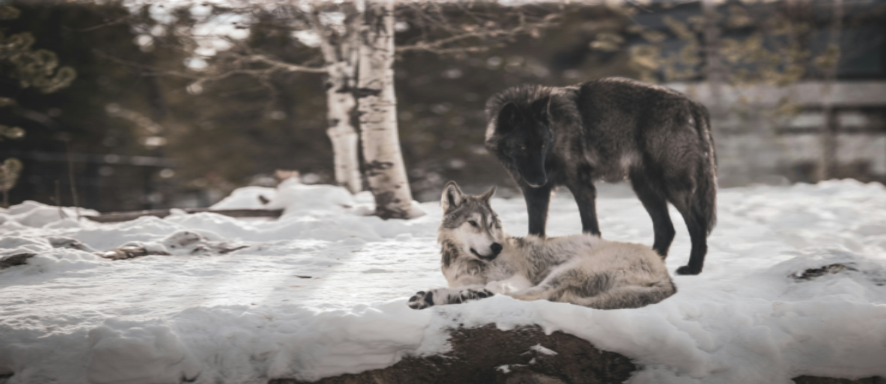}}
    \end{subfigure}
    \vspace{3mm}
    
    \begin{subfigure}[t]{0.30\textwidth}
        \centering
        \frame{\includegraphics[trim=0 0 0 0, clip,width=\textwidth]{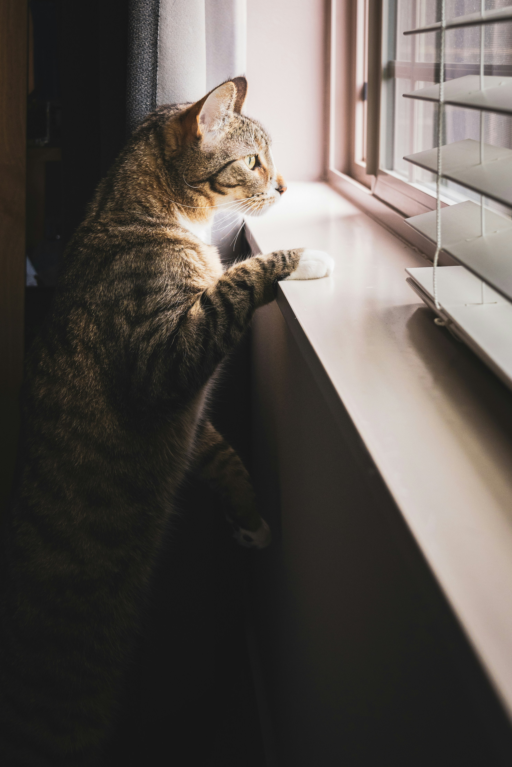}}
    \end{subfigure}
    \hspace{3mm}
    \begin{subfigure}[t]{0.45\textwidth}
        \centering
        \frame{\includegraphics[trim=0 0 0 0, clip,width=\textwidth]{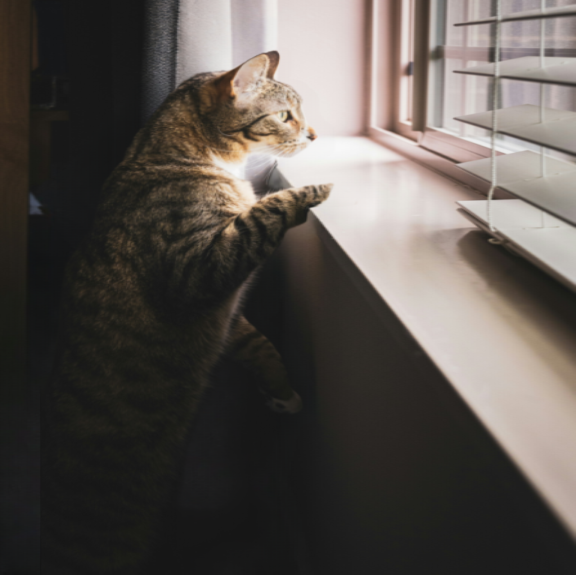}}
    \end{subfigure}
    \vspace{3mm}
    
%    \begin{subfigure}[t]{0.30\textwidth}
%        \centering
%        \frame{\includegraphics[trim=0 0 0 0, clip,width=\textwidth]%{images/supp_affordance/00076_input.png}}
    %\end{subfigure}
   % \hspace{3mm}
   % \begin{subfigure}[t]{0.45\textwidth}
   %     \centering
   %     \frame{\includegraphics[trim=0 0 0 0, clip,width=\textwidth]%{images/supp_affordance/00076_1.50.png}}
    %\end{subfigure}
   \vspace{3mm}
    
    \begin{subfigure}[t]{0.30\textwidth}
        \centering
        \frame{\includegraphics[trim=0 0 0 0, clip,width=\textwidth]{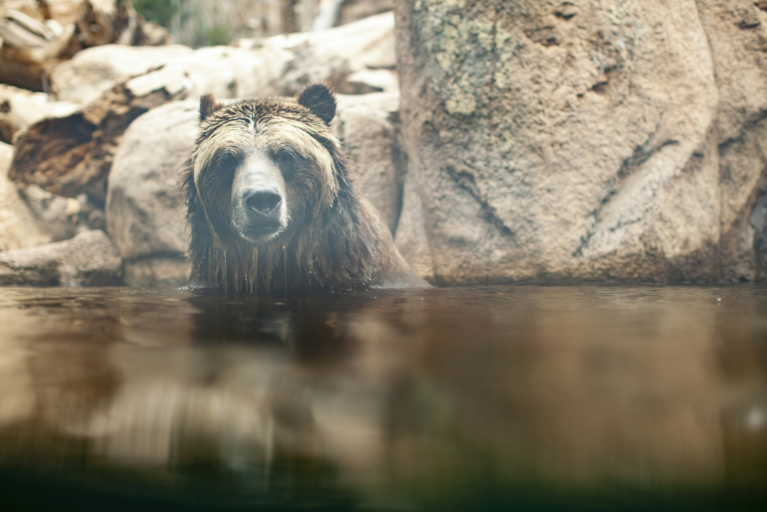}}
        \text{{Input}}
    \end{subfigure}
    \hspace{3mm}
    \begin{subfigure}[t]{0.45\textwidth}
        \centering
        \frame{\includegraphics[trim=0 0 0 0, clip,width=\textwidth]{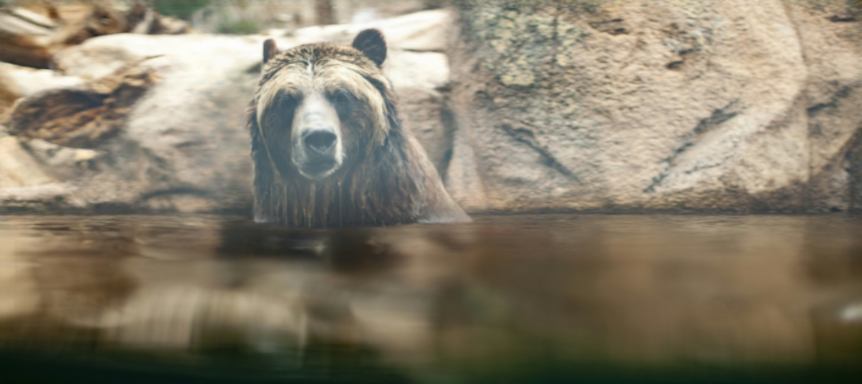}}
        \text{Retargeted} 
    \end{subfigure}
    
    \caption{\textbf{Additional results for affordance-awareness.}
    Our model emerges with an ability to understand the affordance of objects.
    It places the salient object properly with other objects.
    For example, in the ``mushroom'' case, mushrooms are placed near the green moss, similar to the input.
    In the ``wolves'' case, wolves are placed at a similar position as in the input image.
    }
    \label{fig:supp_affordance}
\end{figure}

\begin{figure}[h!]
    \centering
    % \scriptsize
    
    \begin{subfigure}[t]{0.3\textwidth}
        \centering
        \frame{\includegraphics[trim=0 0 0 0, clip,height=0.15\textheight]{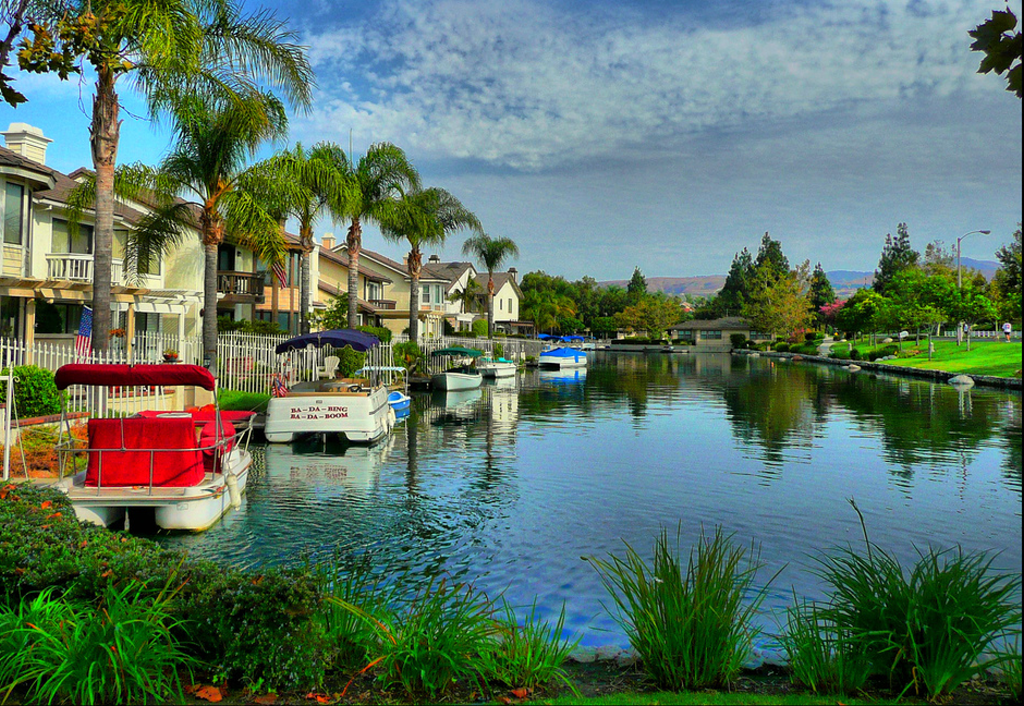}}
        \text{Input }
    \end{subfigure}\hfill
    \begin{subfigure}[t]{0.3\textwidth}
        \centering
        \frame{\includegraphics[trim=0 0 0 0, clip,height=0.15\textheight]{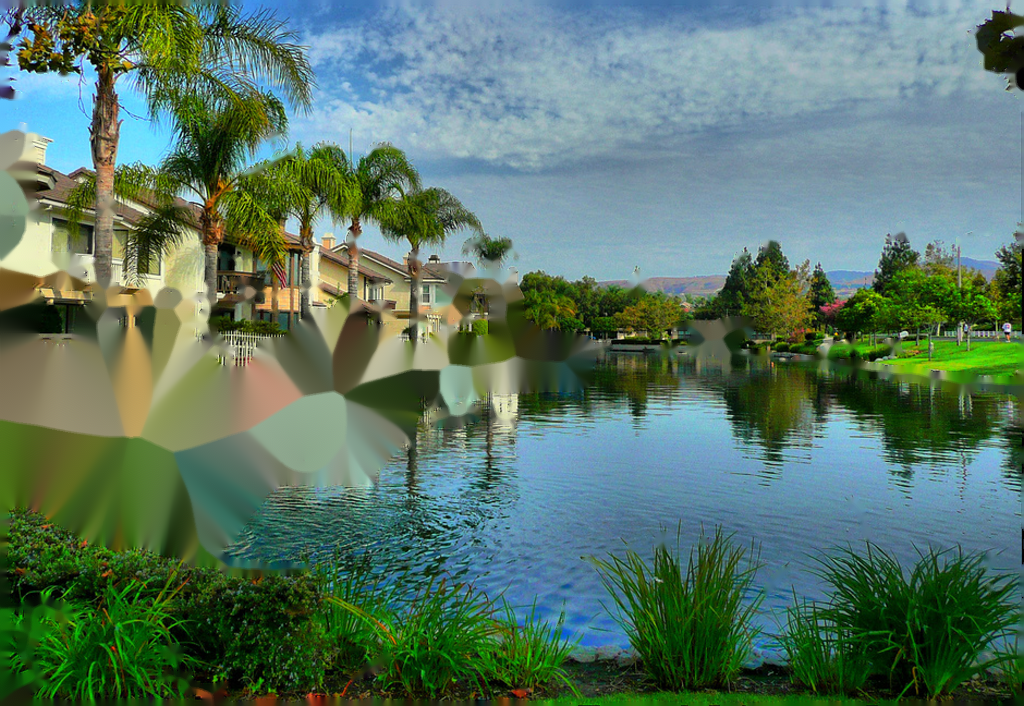}}
        \text{OpenCV}
    \end{subfigure}\hfill
    \begin{subfigure}[t]{0.3\textwidth}
        \centering
        \frame{\includegraphics[trim=0 0 0 0, clip,height=0.15\textheight]{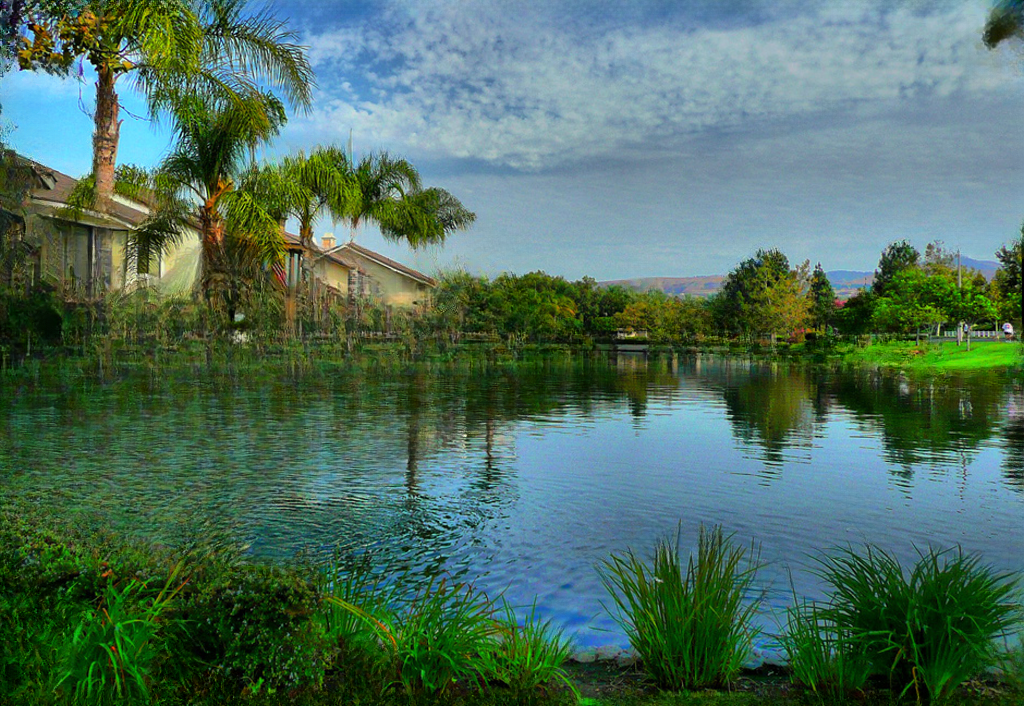}}
        \text{LAMA~\cite{suvorov2022lama}}
    \end{subfigure}

    \caption{\textbf{Why we use LAMA for inpainting~\cite{suvorov2022lama}.}
    We compare LAMA, a state-of-the-art inpaining model, with another off-the-shelf inpainting method from OpenCV. LAMA shows significantly better performance.
    }
    \label{fig:why_lama}
\end{figure}

\begin{figure}[h!]
    \centering
    \begin{subfigure}[t]{0.22\textwidth}
        \centering
        \frame{\includegraphics[trim=0 0 0 0, clip,height=0.23\textheight]{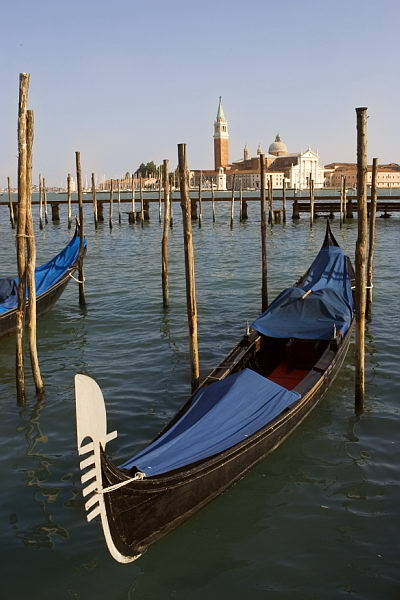}}
        \text{Input}
    \end{subfigure}\hfill
    \begin{subfigure}[t]{0.22\textwidth}
        \centering
        \frame{\includegraphics[trim=0 0 0 0, clip,height=0.23\textheight]{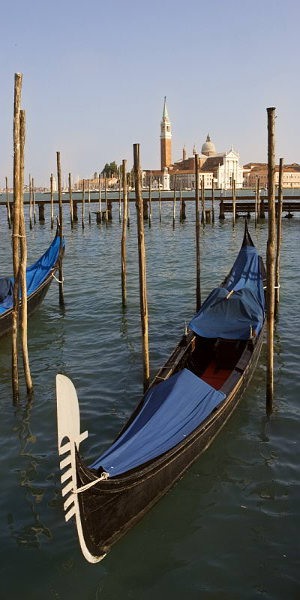}}
        \text{Naively resized}
        \text{MUSIQ$\uparrow$: 53.43}
    \end{subfigure}\hfill
    \begin{subfigure}[t]{0.22\textwidth}
        \centering
        \frame{\includegraphics[trim=0 0 0 0, clip,height=0.23\textheight]{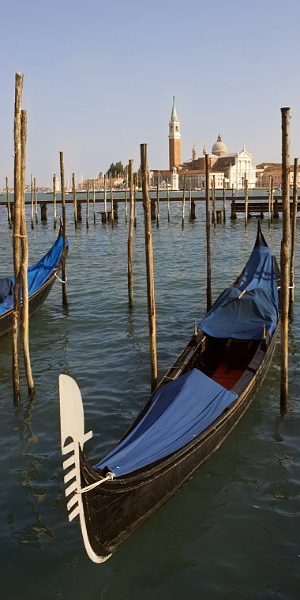}}
        \text{Self-Play-RL}
        \text{MUSIQ$\uparrow$: 53.45}
    \end{subfigure}\hfill
    \begin{subfigure}[t]{0.22\textwidth}
        \centering
        \frame{\includegraphics[trim=0 0 0 0, clip,height=0.23\textheight]{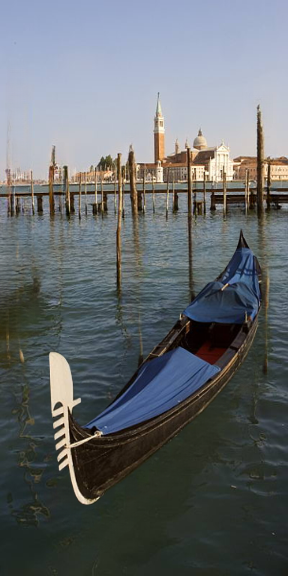}}
        \text{HALO (ours)}
        \text{MUSIQ$\uparrow$: 51.20}
    \end{subfigure}

    \caption{\textbf{Limitation of MUSIQ~\cite{ke2021musiq}.}
     We find MUSIQ itself may \emph{not} be sensitive to the distortions as they are trained with undistorted images. Self-Play-RL~\cite{kajiura2020self} shows a similar result to naively resized output, which has distortions.
     Our result, however, showing less distortions, receives a lower MUSIQ score.
    }
    \label{fig:musiq_limit}
\end{figure}

We show more results to compare with previous approaches:
\begin{itemize}
    \item Generative model overfitting: SINE~\cite{zhang2023sine}, SinDDM~\cite{kulikov2023sinddm}, GPDM~\cite{elnekave2022generating}, GPNN~\cite{granot2022drop}, IDF~\cite{elsner2024retargeting};
    \item Feed-forward approaches: Self-Play-RL~\cite{kajiura2020self}, Cycle-IR~\cite{tan2019cycle}, WSSDCNN~\cite{cho2017wssdcnn}, PR~\cite{shen2024prune};
    \item Drag-style editing: MagicFixup~\cite{alzayer2024magic}, DragonDiffusion~\cite{mou2024dragondiffusion}.
\end{itemize}
The results are shown in Fig.~\ref{fig:supp_qual_comp_1} and Fig.~\ref{fig:supp_qual_comp_3}.
% Fig.~\ref{fig:supp_qual_comp_2}.

\subsubsection{Additional results on the in-the-wild data}
\begin{figure*}[t]    
    \centering
    \captionsetup{type=figure}
    \includegraphics[trim=0 0 0 0, clip,width=\textwidth]{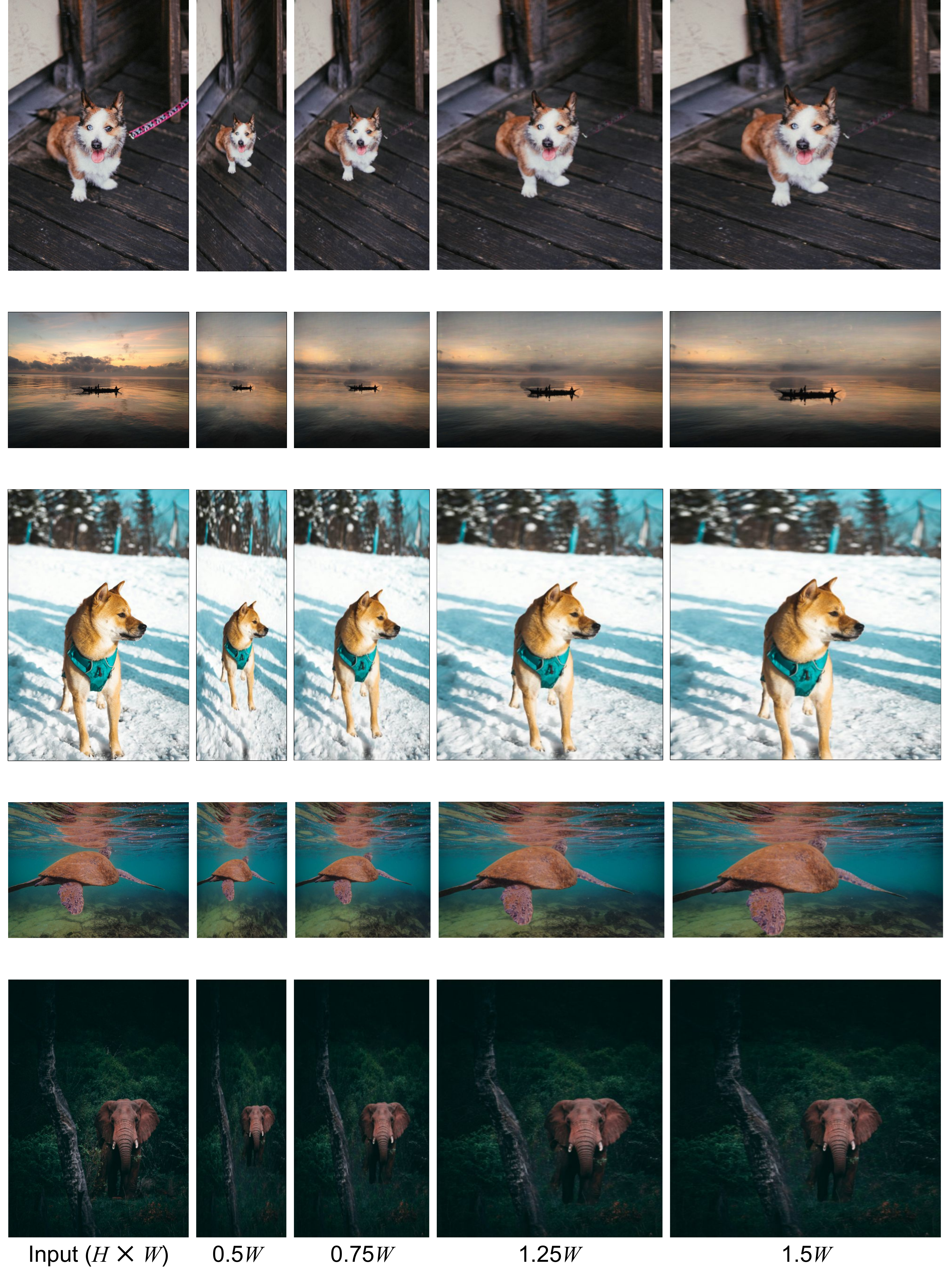}

    \caption{\textbf{Additional qualitative results on the in-the-wild images.} 
    \emph{Without further finetuning}, our model generalizes to the in-the-wild images.
    The input images are from the Unsplash dataset~\cite{unsplash2020unsplash}.
    }
    \label{fig:supp_qual_inwild_1}
\end{figure*}

\begin{figure*}[t]    
    \centering
    \captionsetup{type=figure}
    \includegraphics[trim=0 0 0 0, clip,width=\textwidth]{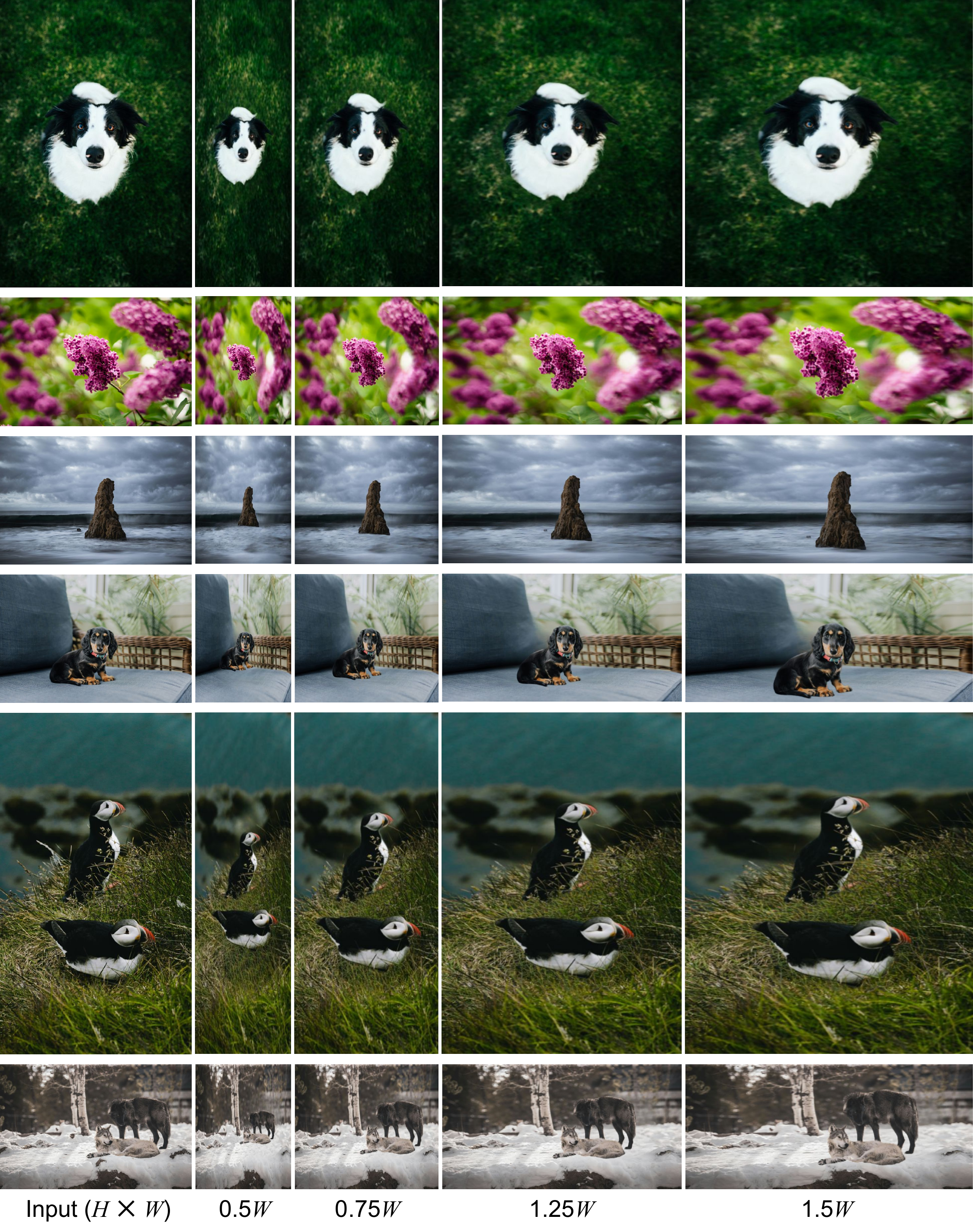}

    \caption{\textbf{Additional qualitative results on the in-the-wild images.} 
    \emph{Without further finetuning}, our model generalizes to the in-the-wild images.
    The input images are from the Unsplash dataset~\cite{unsplash2020unsplash}.
    }
    \label{fig:supp_qual_inwild_3}
\end{figure*}

\begin{figure*}[t]    
    \centering
    \captionsetup{type=figure}
    \includegraphics[trim=0 0 0 0, clip,width=\textwidth]{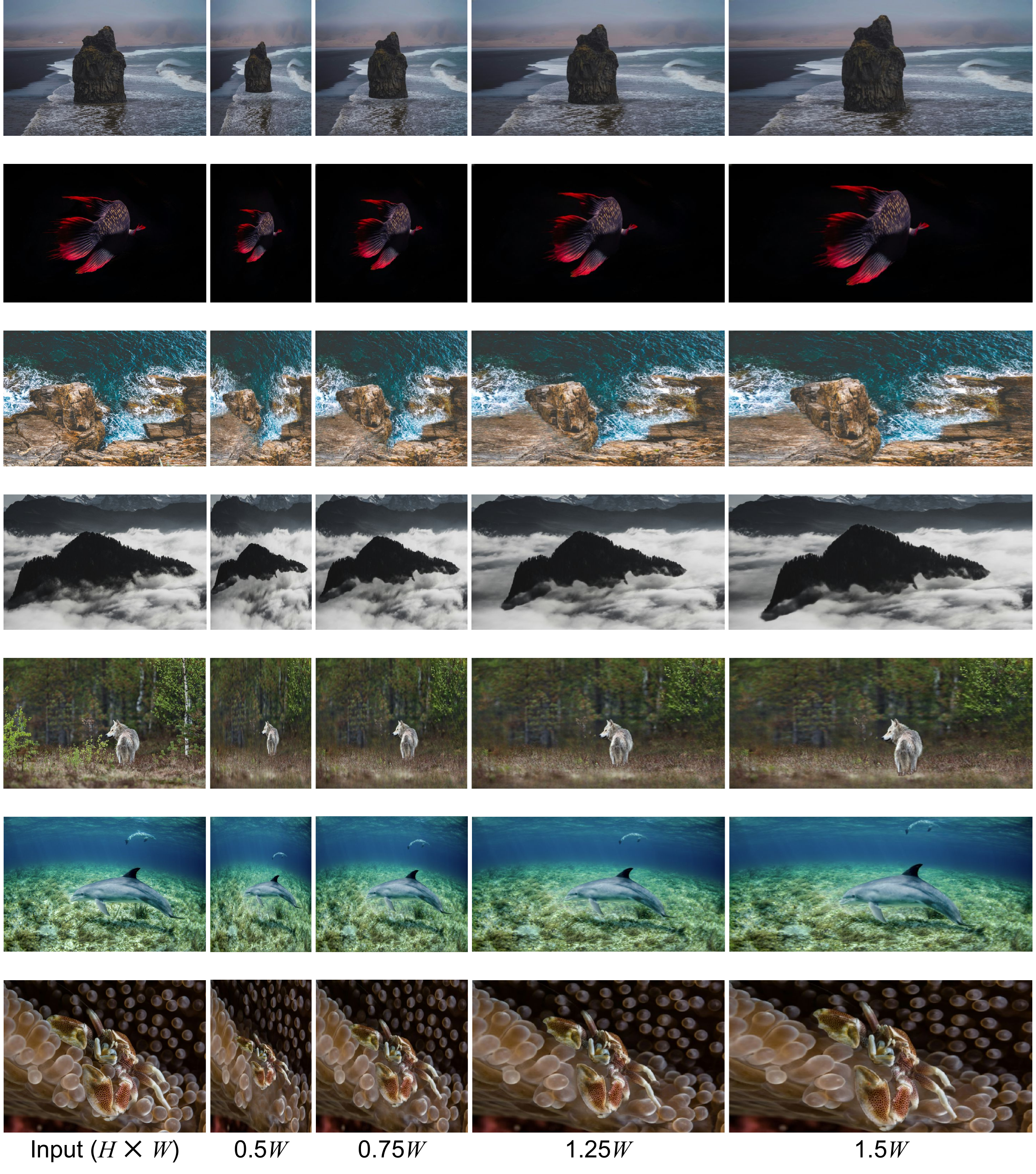}

    \caption{\textbf{Additional qualitative results on the in-the-wild images.} 
    \emph{Without further finetuning}, our model generalizes to the in-the-wild images.
    In ``crab'' case, our method notice the ``affordance'' between the coral and the crab.
    The input images are from the Unsplash dataset~\cite{unsplash2020unsplash}.
    }
    \label{fig:supp_qual_inwild_2}
\end{figure*}

We show more in-the-wild results in Fig.~\ref{fig:supp_qual_inwild_1}, Fig.~\ref{fig:supp_qual_inwild_3} and Fig.~\ref{fig:supp_qual_inwild_2}.


% \subsubsection{Additional results with other no-reference metrics}
% We include additional no-reference metrics in Table~\ref{tab:quant_full_supp}, specifically the learning-based score VILA~\cite{ke2023vila} and the non-learning-based score NIQE~\cite{mittal2012making}. Our HALO method demonstrates competitive performance on these no-reference metrics, achieving the highest average score.

% To compute the average scores, we normalize both VILA and NIQE to percentages. For NIQE, we use $100 - \mathrm{norm(NIQE)}$, as a lower NIQE score indicates better performance.
% \input{tables/tab4_quantitative_comp_full}

% \subsection{LPIPS with our augmentation}
% We additionally show the result using LPIPS with our proposed layout augmentation (Section~\ref{sec:layout_disturb}) in Table~\ref{tab:lpips_aug}.
% With our layout augmentation, the performance of LPIPS is improved, but still worse than the one with DreamSim~\cite{fu2023dreamsim}, as LPIPS is not sensitive to the structure (Fig.~\ref{fig:perceptual_loss_comp}).
% \input{tables/tab6_lpips_aug}

\end{document}